\documentclass{ecsthesis}      
\graphicspath{{Figures/}}   
\usepackage{natbib}            
\usepackage{algorithmicx,algpseudocode,algorithm}
\hypersetup{colorlinks=false}   
\newcommand{\BibTeX}{{\rm B\kern-.05em{\sc i\kern-.025em b}\kern-.08em T\kern-.1667em\lower.7ex\hbox{E}\kern-.125emX}}

\newcounter{address}
\newcommand{\costfunc}{\mathcal{L}}
\DeclareMathOperator{\E}{\mathbb{E}}

\begin{document}
\frontmatter
\title      {Wasserstein Distance Guided Cross-Domain Learning}
\authors    {\texorpdfstring{\href{mailto:js2m17@soton.ac.uk}{Jie Su}}{}}
\supervisor {\texorpdfstring {\href{mailto:apb@ecs.soton.ac.uk}{Adam Prugel-Bennett}}{}}
\addresses  {\groupname\\\deptname\\\univname}
\date       {\today}
\subject    {}
\keywords   {}
\maketitle
\begin{abstract}

Domain adaptation~\citep{pan2010survey} aims to generalise a high-performance learner on target domain (non-labelled data) by leveraging the knowledge from source domain (rich labelled data) which comes from a different but related distribution. Assuming the source and target domains data(e.g. images) come from a joint distribution but follow on different marginal distributions, the domain adaptation work aims to infer the joint distribution from the source and target domain to learn the domain invariant features. Therefore, in this study, I extend the existing state-of-the-art approach to solve the domain adaptation problem. In particular, I propose a new approach to infer the joint distribution of images from different distributions, namely Wasserstein Distance Guided Cross-Domain Learning (WDGCDL). WDGCDL applies the Wasserstein distance~\citep{vallender1974calculation} to estimate the divergence between the source and target distribution which provides good gradient property and promising generalisation bound. Moreover, to tackle the training difficulty of the proposed framework, I propose two different training schemes for stable training. Qualitative results show that this new approach is superior to the existing state-of-the-art methods in the standard domain adaptation benchmark.

\end{abstract}
\tableofcontents 
\listoffigures 
\listoftables
\acknowledgements{First and foremost, I would like to thank my supervisor, Pro. Adam Prugel-Bennett, for his patient, valuable guidance in every stage of the writing of this dissertation. I have been extremely fortunate to have such supervisor who enlightens me not only in this dissertation but also in my future study. I shall extend my thanks to my parents who provide finance and mental support of my master study. My sincere appreciation also goes to my brothers who give me a lot of support on the experiment.}
\originality{
\begin{itemize}
	\item I have read and understood the ECS Academic Integrity information and the University’s Academic Integrity Guidance for Students.
	\item I am aware that failure to act in accordance with the Regulations Governing Academic Integrity may lead to the imposition of penalties which, for the most serious cases, may include termination of programme.
	\item I consent to the University copying and distributing any or all of my work in any form and using third parties (who may be based outside the EU/EEA) to verify whether my work contains plagiarised material, and for quality assurance purposes.
	\item I have acknowledged all sources, and identified any content taken from elsewhere.
	\item I have not used any resources produced by anyone else.
	\item I did all the work myself, or with my allocated group, and have not helped anyone else.
	\item The material in the report is genuine, and I have included all my data/code/designs.
	\item I have not submitted any part of this work for another assessment.
	\item My work did not involve human participants, their cells or data, or animals.
\end{itemize}
}
\mainmatter

\chapter{Introduction}

\section{Research Challenge}
Deep neural networks have been applied in dealing with various machine learning problems and applications. Deep convolutional networks achieve state-of-the-art performance across a variety of tasks such as image classification, object detection and image segmentation. Unfortunately, these impressive performance gains come only when the model is under supervised training with a massive labelled dataset. One of the biggest challenges is that the data labelling process is always arduous and usually involves millions of specialists to do data annotation. To address this problem, there is a strong incentive to establishing effective algorithms to reduce the annotation cost by leveraging existing labelled dataset in related domain areas. However, the challenge of this learning paradigm is the domain shift between different domains which obstructs the adaptation of predictive models from the source domain to the target domain.

\section{Potential Solution}

Domain adaptation~\citep{pan2010survey} methods intend to address the negative effects of domain shift by bridging the source and target domains to learn the domain invariant feature representations without using target labels. In this case, the predictive model learned from the source domain can be applied to the target domain. The previous work of deep domain adaptation methods are conducted under the assumption that the source domain and target domain come from a joint distribution but follow the different marginal distributions. By inferring the joint distribution across different domains (e.g. Mapping both domains into a common feature space), the state-of-the-art algorithms can learn the domain invariant feature representation and perform the domain adaptation. This process is always carried on by minimizing some measure of domain shift such as maximum mean discrepancy. Another way is to reconstruct the target domain from the source domain representation.

Previous studies leverage Maximum Mean Discrepancy~\citep{gretton2007kernel} or KL-divergence~\citep{shlens2014notes} to measure the divergence across different distribution. The potential problem is that when two distributions are distant, it might provide useless gradient information or cause the gradient explosion problem. A more reasonable solution would be to replace the domain discrepancy measure with Wasserstein distance which provides more stable gradient information even two distribution is far away.

Motivated by the shared latent space assumption~\citep{liu2017unsupervised}, I propose a new Wasserstein Distance Guided Cross-Domain Learning (WDGCDL) framework. WDGCDL contains two parts which are the image-to-image translation and the domain adaptation training. The first part aims to infer the joint distribution by learn a shared-latent space across different domains, through reconstructing the target domain from the shared feature representation(and vice versa) to do the image translation task. The second part aims to force the network to interpret the different domains (i.e. translated images and reconstructed images) in the same way to achieve the domain adaptation work.

\section{Contributions}
The main contributions of this study are listed as follows:
\begin{enumerate}
  \item Based on the shared latent space assumption, I applied the Wasserstein distance as a way to measure the domain discrepancy across different distributions. 
  \item I carefully designed two novel training schemes to solve the training difficulty of the proposed framework.
  \item Empirical evidence shows that this approach is simple yet surprisingly powerful and that it outperforms the state-of-the-art models on the MNIST, USPS, and SVHN digital datasets.
\end{enumerate}

\section{Dissertation Structure}
\textbf{Chapter 1} describes the motivation behind the work proposed in this study, and highlights the main contributions of the framework.

\textbf{Chapter 2} presents the basic background knowledge that helps the understanding of the wholes study.

\textbf{Chapter 3} explores and summarizes related work on domain adaptation.

\textbf{Chapter 4} describes the details of the proposed framework, and the novel training algorithms that designed for stable training.

\textbf{Chapter 5} presents the evaluation of the proposed framework, and explains the results.

\textbf{Chapter 6} summarises and provides a conclusion of the work presented in this study and proposes directions for future studies in the related areas.

\section{Notation}

In this study, the lower-case characters~(e.g. $x,y$ and etc.) indicate the small vector terms which comes from a big vector space (e.g. $X,Y$) or some probability distribution space (e.g. $P_r,Q_{\theta}$). Moreover,	equation \ref{eq:dsitance} is used to describe the distance measurement.
\begin{equation}
\label{eq:dsitance}
	D_{name\ of\ distance}(X,Y) = ||\cdot||
\end{equation}
where $||\cdot||$ can be any function used to measure the distance. 

$\costfunc_{name}$ is used to represent the cost function or objective function. The greek letter (e.g. $\delta,\theta$)is used to describe the network parameter or put it as the subscript of distribution. 

\chapter{Background}
This section will introduces some important basic knowledges which will go through the whole study such as different distance measurements, Auto-Encoders and Generative Adversarial Network. 
\section{Distance Measurement}
\subsection{Minkowski Distance}
The Minkowski distance~\citep{benz2000lorentz} is a metric induced by the $L_p$ norm, that is, the metric in which the distance between two vectors is the norm of their difference. A norm is a function that can describe the length or size of a vector. The $L_p$ norm of a vector $x$ with $i$ components is defined as:
\begin{equation}
||x||_p=(\sum_{i=1}^{n}|x_i|^p)^{\frac{1}{p}}
\end{equation} 
Assuming there are two vectors $X=(x_1,...,x_n)$ and $Y=(y_1,...,y_n)\in \mathbb{R}^n$, then the Minkowski distance family can be formulated as:
\begin{equation}
	D_{Minkowski}(X,Y)=(\sum_{i=1}^{n}|x_i-y_i|^p)^{\frac{1}{p}}
\end{equation}
There are two generalization distances in the Minkowski distance family that are widely used in machine learning area which called Manhattan distance~\citep{black1998dictionary} and Euclidean distance~\citep{liberti2014euclidean}.
\subsection{Manhattan Distance}
In an $n-$dimension real vector space with fixed Cartesian coordinate system, two points can be connected by a straight line. The sum of the line's projection onto the coordinate axes is the Manhattan distance. It is the situation when the $L_p$ norm in Minkowski distance equals $L_1$, so that it also called $L_1$ distance which can be formulated as:
\begin{equation}
	D_{Manhattan}(X,Y)=\sum_{i=1}^{n}|x_i-y_i|
\end{equation}
\subsection{Euclidean Distance}
Different with the Manhattan distance, the Euclidean distance~\citep{liberti2014euclidean}($L_2$ distance) measures the straight line distance between two points in the vector space. It is also a generalization of the Minkowski distance when $L_p\rightarrow L_2$ which can be formulated as:
\begin{equation}
	D_{Euclidean}(X,Y)=\sqrt{\sum_{i=1}^{n}|x_i-y_i|^2}
\end{equation}
\subsection{Kullback-Leibler Divergence}
In mathematical statistics, KL-Divergence~\citep{shlens2014notes}(or relative entropy) is a way to measure the distance or the divergence of one probability distribution and another probability distribution. Assuming there are two probability distributions $P_r$ and $Q_{\theta}$ over a set $\mathcal{X}$, the KL-Divergence from $Q_{\theta}$ to $P_r$ can be formulated as:
\begin{equation}
	D_{KL}(P_r||Q_{\theta}) = -\sum_{\mathcal{X}}P_r(x)\log \frac{Q_{\theta}(x)}{P_r(x)}
\end{equation} 
\subsection{Maximum Mean Discrepancy}
Maximum Mean Discrepancy~\citep{gretton2007kernel}(MMD) is another measurement of the distance between probability distributions which has found numerous applications in machine learning. MMD is defined by the idea of representing probability distribution distance as the distance between mean embeddings of features. Assuming there is a feature mapping function $f: \mathcal{X}\rightarrow \mathcal{H}$, where $\mathcal{H}$ is a reproducing kernel Hilbert space. Then, the MMD can be formulated as:
\begin{equation}
	D_{MMD}(P_r,Q_{\theta}) = ||\mathbb{E}_{X\sim P_r}[f(X)]-\mathbb{E}_{Y\sim Q_{\theta}}[f(Y)]||_{\mathcal{H}}
\end{equation}
\subsection{Wasserstein Distance}
For discrete probability distributions, the Wasserstein distance~\citep{vallender1974calculation}(or Kantorovich-Monge-Rubinstein metric) is also descriptively called the earth mover's distance(EMD). Imagining the probability distributions are the different heaps of a certain amount of earth, the EMD is the minimal cost to transferring one heap to the other.  Assuming there are two distributions $P_r$ and $Q_{\theta}$ with $l$ possible states $x$ or $y$. Then, by defining the transport plan as $\gamma(x,y)$, the EMD can be formulated as the multiplication of transport plan with the Euclidean distance between $x$ and $y$.
\begin{equation}
	D_{EMD}(P_r,Q_{\theta}) = \underset{\gamma\in\Pi}{\inf}\underset{x,y}{\sum}||x-y||\gamma(x,y) = \underset{\gamma\in\Pi}{\inf}\mathbb{E}_{(x,y)\sim\gamma}||x-y||
\end{equation}
where the expression $\inf$ means infimum or greatest lower bound.

For the continuous probability distributions, the Wasserstein distance can be viewed as the situation when discrete distributions expand to many infinite states. Then it can be formulated as:
\begin{equation}
	D_{Wasserstein}(P_r,Q_{\theta}) = \underset{\gamma\in\Pi}{\inf}\underset{x,y}{\iint}||x-y||\gamma(x,y)dxdy = \underset{\gamma\in\Pi}{\inf}\mathbb{E}_{(x,y)\sim\gamma}||x-y||
\end{equation}
\section{Convolutional Neural Networks}

In the last two decades, with the appearance of convolutional neural network(CNN)~\citep{lecun1998gradient}, the computer vision has witnessed a significant development.
\begin{figure}[h]
	\centering
		\includegraphics[width=1.0\textwidth]{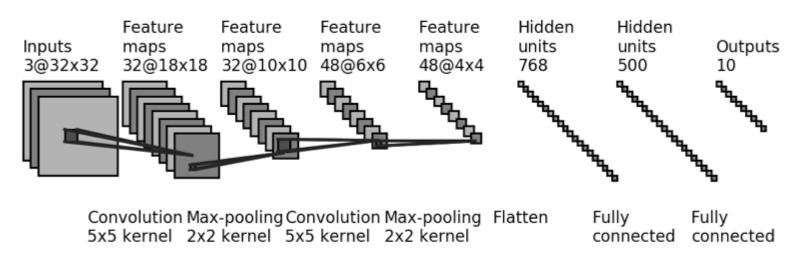}
	\caption[Convolutional Neural Network Architecture]{Convolutional Neural Network Architecture\citep{ding2016automatic}}
	\label{fig:cnn}
\end{figure}

Figure \ref{fig:cnn} depict the workflow of convolutional neural network when inputting the RGB image. This network includes two convolutional layers, two sub-sampling layers, and also, three fully-connected layers. The appropriate selection of kernels and filters bring us excellent properties(i.e. sparse connectivity and shared weights)of this network. For example, in figure \ref{fig:cnn}, the kernel size is set as $5\times 5\times 3$, and it can perfectly extract the features from the images we has input previously. Besides, a good choice of stride (ie.e stride=2,3,...n) would result in less weights in some important margins as well as reducing the risk of over-fitting.

\begin{figure}[h]
  \centering
  \begin{minipage}[t]{0.41\textwidth}
    \includegraphics[width=\textwidth]{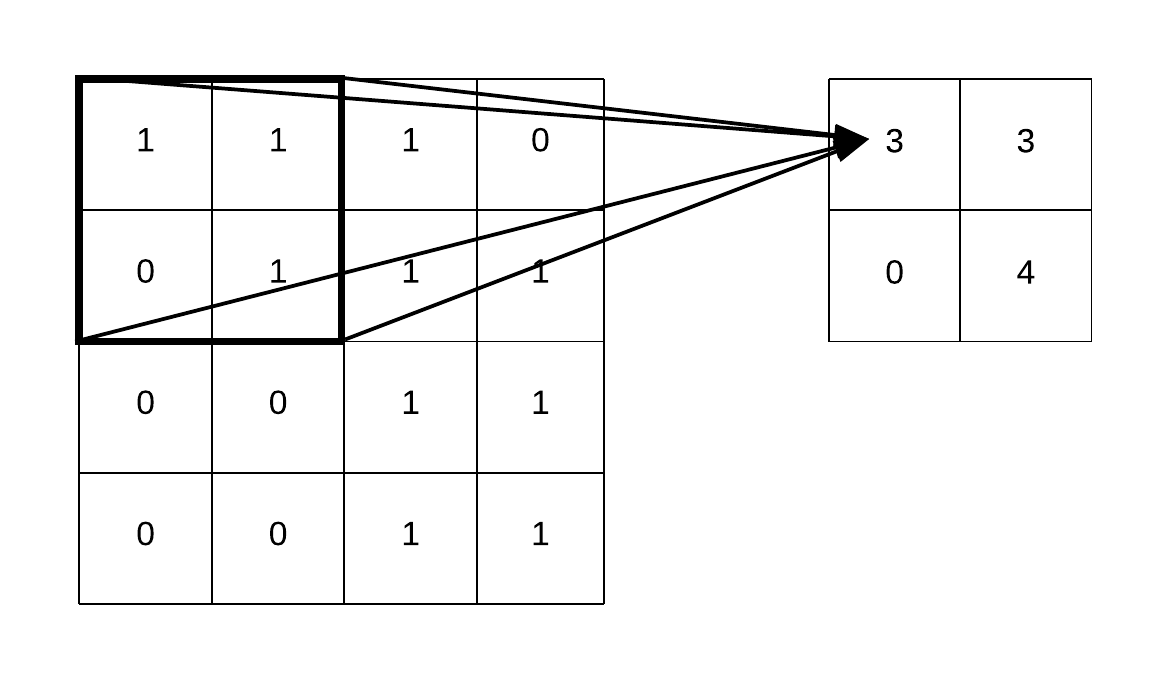}
    \caption{Feature Extract Process with stride=$2$}
    \label{fig:conv}
  \end{minipage}
~
  \begin{minipage}[t]{0.45\textwidth}
    \includegraphics[width=\textwidth]{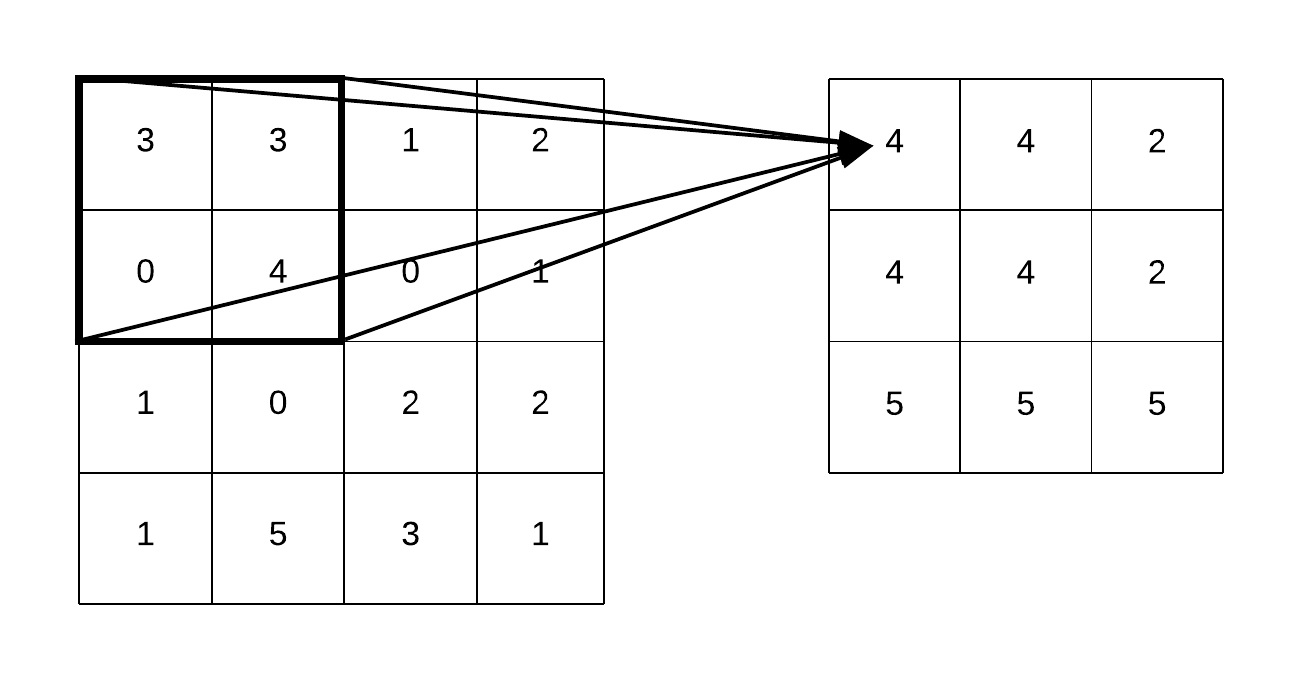}
    \caption{Max Pooling Process}
    \label{fig:pool}
  \end{minipage}
\end{figure}

After the convolutional layers, there always exist two layers of sub-sampling, for example, the max pooling layer, the mean pooling layer and the min pooling layer. The reason why we need the sub-sampling layers is that they could effectively reduce the scale of feature maps, and at the same time, keep the invariance~\citep{scherer2010evaluation} of the space. The max pooling layers are widely used to obtain the largest amount of moving slices(i.e., figure \ref{fig:pool}).

Finally, the fully connected layer at the end of network will attach a softmax function(if that is a multi-class task) to output the classification result.

\section{Auto-Encoders}
\begin{figure}[h]
	\centering
		\includegraphics[width=1.0\textwidth]{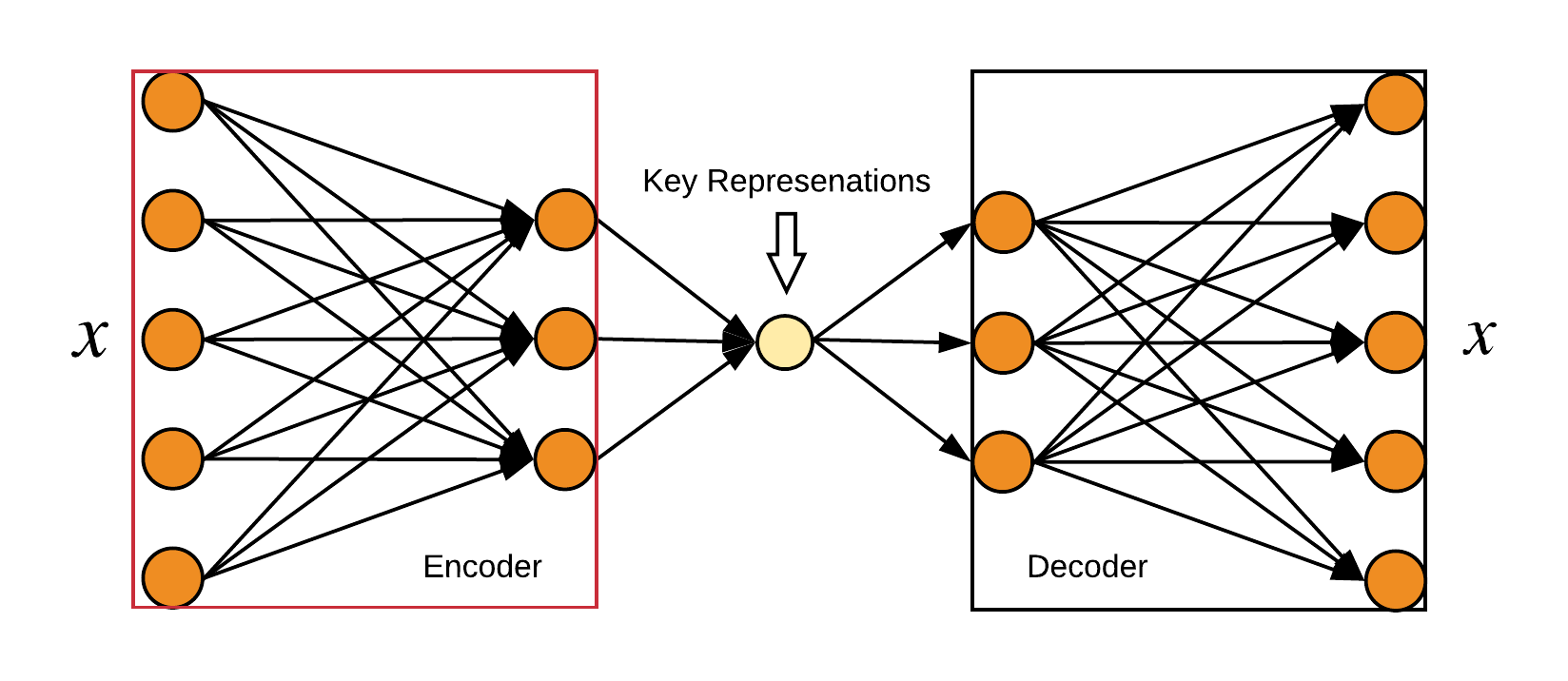}
	\caption[Auto-Encoders Architecture]{Auto-Encoders Architecture}
	\label{fig:autoencoders}
\end{figure}

Auto-Encoders(AEs)\citep{rumelhart1985learning} are simple learning circuit which uses the input as the teacher of the framework. It aims to transfer the input to the output with the least possible amount of distortion and learn the key representation from high dimension input data(e.g. Images). Define the input $x$ and the autoencoder parameters $h_{w,b}$, then the autoencoder aims to learn a function that:
\begin{equation}
	h_{w,b}(x)\approx x
\end{equation}
The auto-encoders architecture(Figure \ref{fig:autoencoders}) contains two parts: Encoder and Decoder. The encoder compresses the input data from high dimension space to a low dimension space while the decoder aims to recover the input data from the low dimension space. The model will be trained by minimizing the $L_1$ distance between the origin input and output:
\begin{equation}
	\costfunc_{w,b} = ||h_{w,b}(x) - x||_1
\end{equation}

Recently, the combination of Auto-Encoders and CNN~(i.e. CAEs~\citep{masci2011stacked}) have been applied to learn good representation from the images by setting CNN as the encoder and decoder parts. 
\subsection{Variational Auto-Encoders}
Variational Auto-Encoders(VAEs)~\citep{kingma2013auto} is an upgrade version of Auto-Encoders. Instead of learning a "compressed representation" of input(e.g. images, text), VAEs learn the parameters of a distribution representation or the latent variable of the data. As it learns to model data, VAEs can generate new input data samples by sampling from the distribution.
\begin{figure}[h]
	\centering
		\includegraphics[width=1\textwidth]{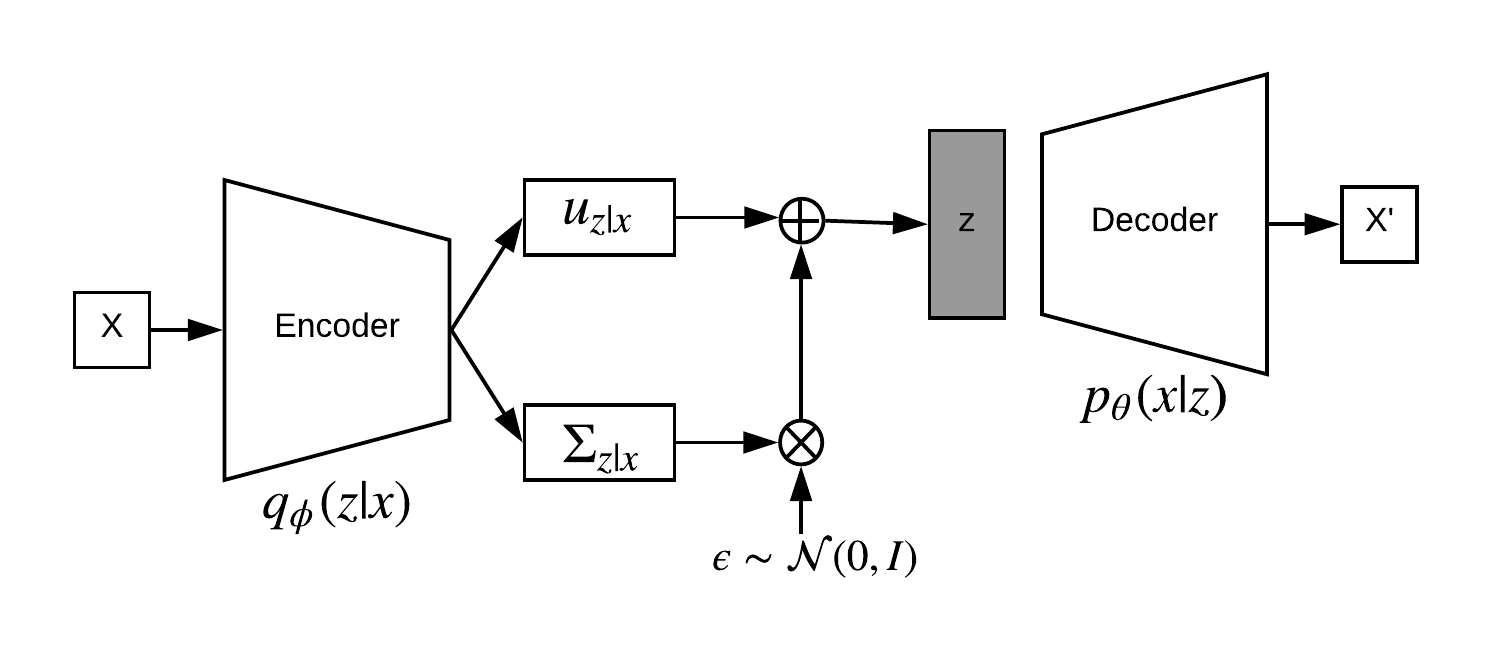}
	\caption[Variational Auto-Encoder Architecture]{Variational Auto-Encoder Architecture}
	\label{fig:vae}
\end{figure}

Figure \ref{fig:vae} illustrates a simple VAEs framework which inherits the encoder-decoder structure of the AEs. The encoder aims to generate a latent variable $z$ by generating $u_{z|x}$ and $\Sigma_{z|x}$ from the data. Then, the decoder $p_{\theta}(x|z)$ will reconstructs the input data from the latent variable. 

Assuming the latent variable follows a prior distribution $p_{\theta}(z)$(e.g. simple Gaussian distribution), the VAEs will be trained by forcing the encoder probability distribution $q_{\phi}(z|X=x)$ to match the $p_{\theta}(z)$. In this case, the objective of VAEs framework~(also called variational upper bound) will consist two parts: The KL-Divergence distance between $q_{\phi}(z|X=x)$ and $p_{\theta}(z)$, and the log-likelihood of the reconstructed input data.
\begin{equation}
	\costfunc_{VAEs}=-\text{KL}(q_{\phi}(z|x)|p_{\theta}(z))+\mathbb{E}[\log p_{\theta}(x|z)]
\end{equation}

\textbf{The Reparameterization Trick:} This trick indicates that the complicated posterior would follow a Gaussian distribution with proximately diagonal covariance and its structure would be like $\log q_{\phi}(z|x)=\log \mathbb{N}(z;\mu,\sigma^2I)$, while the prior distribution about latent variables would center around an isotropic multivariate Gaussian distribution, $p_{\theta}(z)=\mathbb{N}(z;0,I)$. 

Since both $p_{\theta}(z)$ and $q_{\phi}(z|x)$ of this model are Gaussian, the resulting estimator for this model and datapoint x is:
\begin{equation}
	\costfunc(\theta,\phi;x)\simeq\frac{1}{2}\sum_{j=1}^{J}(1+log\sigma_j^2-\mu_j^2-\sigma_j^2)+\frac{1}{L}\sum_{l=1}^{L}log p_{\theta}(x|z^l)
\end{equation}
where $z^l=\mu+\sigma\odot\epsilon^l$ and $\epsilon^l\sim\mathbb{N}(0,I)$. An optimizer can then be used to maximize the $\costfunc(\theta,\phi;x)$.

\subsection{Adversarial Auto-Encoders}
\begin{figure}[!htp]
\centering
\includegraphics[width=0.7\textwidth]{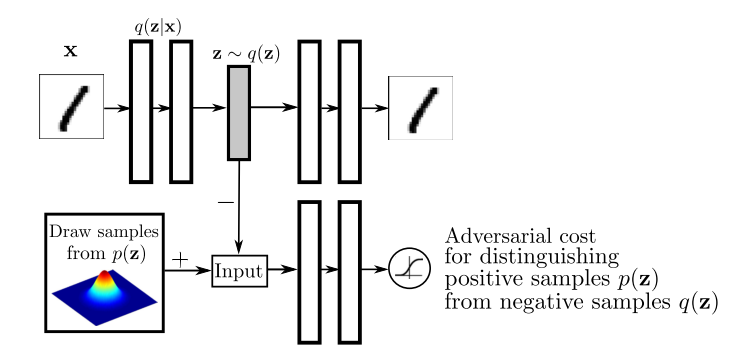}
\caption[Architecture of an adversarial autoencoder~\citep{makhzani2015adversarial}]{Architecture of an adversarial autoencoder~\citep{makhzani2015adversarial}. The top row is a standard autoencoder that reconstructs an image $x$ from a latent code $z$. The bottom row diagrams a second network trained to discriminatively predict whether a sample arises from the hidden code of the autoencoder or from a sampled distribution specified by the user.}
\label{fig:aae}
\end{figure}

Adversarial Auto-Encoders(AAEs)~\citep{makhzani2015adversarial} first introduces the adversarial training scheme to the auto-encoder structure. Different from the VAEs, AAEs replace the KL-divergence regularisation term to an adversarial procedure to impose an arbitrary prior on the latent code. Therefore, the net will generate latent code that is indistinguishable with the prior latent code. Figure \ref{fig:aae} and Equation \ref{eq:aaes} illustrate the basic framework structure of AAEs and the object function of AAEs.
\begin{equation}
\label{eq:aaes}
	\costfunc_{AAEs}= Adv(q_{\phi}(z|x)|p_{\theta}(z)) + \mathbb{E}[\log p_{\theta}(x|z)]
\end{equation}
where $Adv(\cdot)$ indicates the adversarial loss. 

The training of AAEs contain two phases: \emph{reconstruction} and the \emph{regularization} phases. In the reconstruction phase, the autoencoder will update the encoder and the decoder to minimize the reconstruction error of the inputs. In the regularization phase, the adversarial network first updates its discriminative network; then updates its generator (encoder).

\subsection{Wasserstein Auto-Encoders}
Wasserstein Auto-Encoders(WAEs)~\citep{tolstikhin2017wasserstein} have a similar encoder-decoder architecture as the VAEs but apply a penalized form of Wasserstein distance between the model distribution and the target distribution to train the network. Instead of forcing the encoder probability distribution $q_{\phi}(z|X=x)$ for every input data to match the prior probability distribution $p_{\theta}(z)$, WAEs force the exception of encoder probability distribution $\int q_{\phi}(z|x)dp_x$ to match the prior probability distribution which reduces the interaction of the latent variable between different samples.  

By introducing the Wasserstein distance or Optimal Transport(OT) cost to measure the discrepancy between different distributions. WAE aims to find out a conditional distribution $Q(Z|X)$ such that its $Z$ marginal distribution is identical to the prior distribution(i.e. $P_Z=Q_Z(Z):=\mathbb{E}_{X\sim P_X}[Q(Z|X)]$). In this case, for a decoder $P_G(X|Z)$ with function $G$ that map $Z\rightarrow X$, the objective of WAE can be written as:
\begin{equation}
	\underset{Q:Q_Z=P_Z}{inf}\mathbb{E}_{P_X}\mathbb{E}_{Q(Z|X)}[c(X,G(Z))]
\end{equation}
where $c(\cdot )$ could be any cost function(e.g. $L_1$ distance, $L_2$ distance).

Additionally, in order to implement the numerical solution, WAE relaxes the constrain on $Q_Z$ by adding a penalty to force the $Q_Z\simeq P_Z$. The final objective can be written as:
\begin{equation}
	D_{WAE}(P_X,P_G):= \underset{Q(Z|X)\in \mathcal{Q}}{inf}\mathbb{E}_{P_X}\mathbb{E}_{Q(Z|X)}[c(X,G(Z))]+\lambda\cdot D_Z(Q_Z,P_Z)
\end{equation}
where $D_Z$ is an arbitrary divergence $Q_Z$ and $P_Z$.

\subsection{A Short tour from AEs to WAEs} 

\textbf{Connecting VAEs to AEs:} Variational auto-encoders(VAEs)~\citep{kingma2013auto} aim to learn a distribution representation of data instead of key representation in AEs. VAEs minimize an upper bound on the negative log-likelihood or, equivalently, on the KL-divergence term. The negative log-likelihood term can be viewed as the reconstruction term of an AE, and KL-divergence term can be viewed as regularization terms which minimize the cross-entropy between the aggregated posterior $q_{\phi}(z)$ and the prior $p_{\theta}(z)$. 

\textbf{Connecting AAEs to VAEs:} Adversarial autoencoders(AAEs)~\citep{makhzani2015adversarial} uses the proposed generative adversarial networks to perform variational inference, it replaces the KL-divergence terms with an adversarial training procedure
that encourages $q_{\phi}(z)$ to match with the whole distribution of $p_{\theta}(z)$.

%

\textbf{Connecting WAEs to VAEs:} Wasserstein Autoencoders(WAEs)~\citep{tolstikhin2017wasserstein} switch the focus from KL-divergence term on VAEs to the optimal transport distance~(i.e. 1-Wasserstein distance). WAEs share most nice properties of VAEs~(e.g. nice latent manifold structure, auto-encoder structure) while providing better quality samples. Moreover, WAEs allow adversary-free training which leading to stable training.

\textbf{Connecting	 WAEs to AAEs:} The adversary version of Wasserstein Autoencoders~(WAEs) replace the adversary-free arbitrary divergence term~(MMD) with an adversarial training procedure. This forces the encoder probability distribution to match to the arbitrary probability distribution. Moreover, WAEs can use any cost function $c(x,y)$ in the input space while AAEs use the second order Wasserstein distance to measure the distance between two probability distributions~(i.e. When the WAEs use cost function $c(x,y)=||x-y||_2^2$, it is equal to AAEs).

\section{Generative Adversarial Networks}
\begin{figure}[h]
	\centering
		\includegraphics[width=1.0\textwidth]{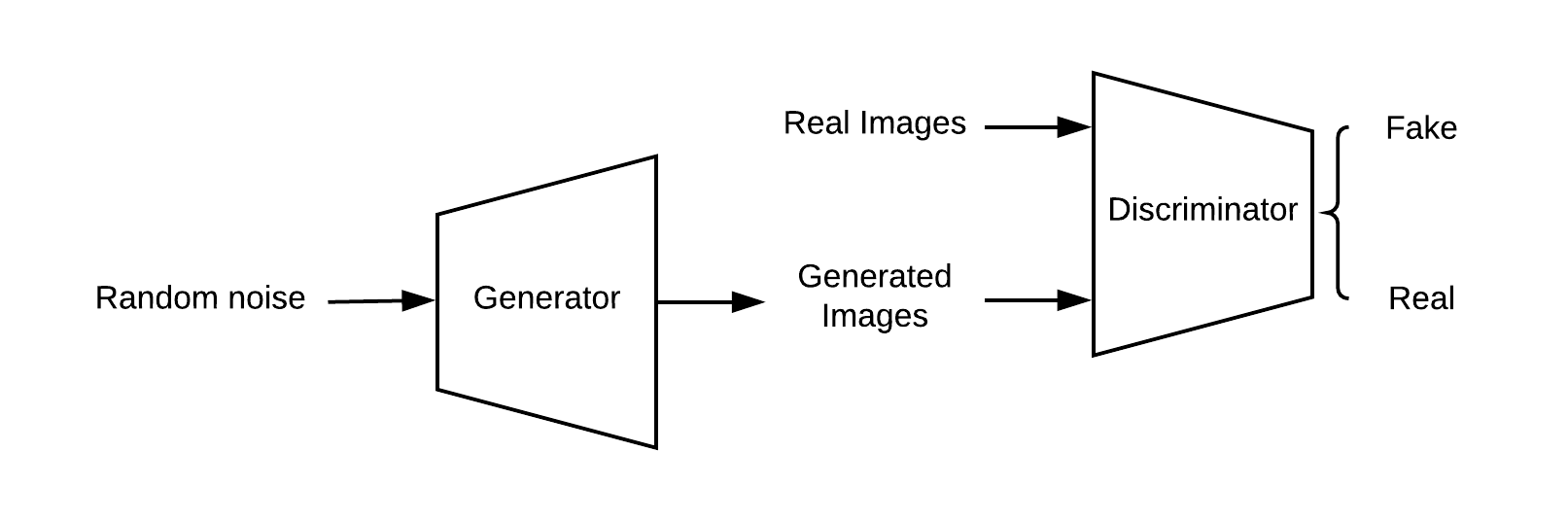}
	\caption[Generative Adversarial Network Architecture]{Generative Adversarial Network Architecture}
	\label{fig:gan}
\end{figure}

Generative Adversarial Networks(GANs) \citep{goodfellow2014generative} were proposed in 2014 as a generative model with adversarial architecture. Figure \ref{fig:gan} shows the basic network architecture of GANs which contain a generator and a discriminator. 
In the adversarial game, the generator aims to generate realistic samples from random noise while the discriminator aims to distinguish the generated samples from real data. Finally, the best situation is that the competition will drive the generated samples to be indistinguishable from the real data. 

Assuming there some real images $X=\{x_1,...x_n\}$, a Generator $G_{\theta}$, a Discriminator $D_{\phi}$ and some random noise $z$. Then the objective of GANs can be seen as a minimax game:
\begin{equation}
	\underset{\phi}{max}~\underset{\theta}{min}[\mathbb{E}_{x\sim p_{data}}\log D_{\phi}(x)+\mathbb{E}_{z\sim p(z)}\log(1-D_{\phi}(G_{\theta}(z))) ]
\end{equation}
In this minimax game, the discriminator aims to maximize the objective such that $D_{\phi}(x)$ is close to 1(real) and $D_{\phi}(G_{\theta}(z))$ is close to 0(fake). The generator aims to minimise the objective such that $D_{\phi}(G_{\theta}(z))$ is close to 1~(i.e. Fool the discriminator by generating real-like data). More precisely, the objective of discriminator and generator can be written as:
\begin{align}
Discriminator&:\ \underset{\phi}{max}~[\mathbb{E}_{x\sim p_{data}}\log D_{\phi}(x)+\mathbb{E}_{z\sim p(z)}\log(1-D_{\phi}(G_{\theta}(z))) ]\\
Generator&:\ ~\underset{\theta}{min}~\mathbb{E}_{z\sim p(z)}\log(1-D_{\phi}(G_{\theta}(z)))
\end{align}

Recently, the combination of VAEs and GANs~(e.g. VAE-GANs~\citep{larsen2016autoencoding}) provide higher image quality than the pure VAEs framework.

\chapter{Related Work}
The challenge of domain adaptation(DA) is the domain shift or dataset bias\citep{gretton2009covariate}. The potential solution to solve this problem is to fine-tune the pre-trained model and apply it to the target task. However, for the non-label dataset, this solution is no longer useful. This chapter will introduce some related works of domain adaptation(DA) tasks from traditional architecture to the state-of-the-art ones.

\section{Traditional Domain adaptation Methods}
\cite{fernando2014subspace} proposed Subspace Alignment(SA) to learn an alignment between the source and target subspace(Subspace were achieved by PCA) where PCA dimension were selected by minimising the divergence between the subspace. Similarly, Correlation Alignment(CORAL)~\citep{sun2016deep} were proposed to reduce the domain shift by using the second-order statistics of the source and target distributions. Instead of aligning the features, feature transformation methods(e.g. TCA~\citep{pan2011domain}) aims to find a projection of data into a latent space where the divergence across different distribution is decreased.

The traditional domain adaptation methods aims to search a latent space(e.g. kernel space, PCA subspace) where the domain discrepancy across different domains is small.

\section{Deep Domain Adaptation Methods}
\textbf{Shallow method with Deep feature:} with the great success of deep convolutional architectures, some naive methods began to use deep convolutional network as the feature extract and applied traditional methods such as Subspace Alignment~\citep{fernando2014subspace}, Correlation Alignment~\citep{sun2016deep} and Feature transformation~\citep{pan2011domain} on it. 

\textbf{Fine-tuning deep CNN architectures:} Another option which is widely used is to fine-tune the pre-trained network to fit the new data on the target domain~\citep{oquab2014learning}. However, the fine-tuning solution is only available when target domain contains some labels.

\textbf{Discrepancy-based methods:} Inspired by the shallow feature space transformation solution. \citep{tzeng2014deep} proposed Deep Domain Confusion~(DDC) net by applying simple linear MMD between the source and the target to select the layers that need to be fine-tuned. Instead of using simple linear MMD, \cite{long2015learning} proposed to use multiple kernels variant of maximum mean discrepancies~(MK-MMD) defined between several layers. The linear combination of those kernels provide a more accurate distance measurement which improve the feature transferability in task-specie layers.

\textbf{Adversarial discriminative models:} By applying the adversarial object to the model, these methods aim to encourage domain confusion to a domain discriminator. \cite{ganin2016domain} proposed Domain-Adversarial Training of Neural Networks(DANN) with a gradient reversal layer that aims to confuse the domain classifier by forcing the extractor  to extract the invariant features from different domains. Adversarial Discriminative Domain Adaptation(ADDA)~\citep{tzeng2017adversarial} applies an inverted label GAN loss that separates the objective into two parts. This ensures that feature extractor can learn the domain specific features.

\textbf{Adversarial generative models:} The Adversarial generative models first combine the discriminator with a generator component(i.e. GANs). Coupled Generative Adversarial Networks(CoGANs)~\citep{liu2016coupled} introduces a couple GANs structure for learning a joint distribution~(or domain invariant feature space) under the weight sharing constraint.~\cite{bousmalis2017unsupervised} proposed to use GAN to generates a source domain image that looks as if it were extracted from the target domain, and then apply the source classifier to the target domain. 

\textbf{Data reconstruction(encoder-decoder) based Methods:}  Different from the above methods, Deep reconstruction network~\citep{ghifary2016deep} jointly solves the source classification problem as well as unsupervised target data reconstruction problem. The data reconstruction can be seen as a task that supports the adaptation of the label prediction. The Domain Separation Networks (DSN)~\citep{bousmalis2016domain} proposed a private subspace which contains domain specific properties. By integrating a reconstruction loss with a shared decoder, this framework can reconstruct input sample from domain specific and source representation.

\textbf{Adversarial Reconstruction Methods:} \cite{liu2017unsupervised} proposed UNsupervised Image-to-image Translation(UNIT) framework with a elegant shared latent space assumption. By applying adversarial training to learn the image reconstruction, image translation and classification tasks, the Couple VAE-GAN framework achieve the excellent results on benchmark dataset. 
\chapter{Methodology}

\section{Unsupervised Image-to-image Translation Framework}
As mentioned in the previous chapter, UNIT~\citep{liu2017unsupervised} achieved state-of-the-art accuracies in the domain adaptation tasks by inferring a joint distribution from marginal distributions. UNIT introduces a elegant shared-latent space assumption which gives more interpretation of the framework and makes it easier to implement. In this study, the proposed framework will be based on the same assumption.
\subsection{Shared-Latent Space Assumption}
How to use the marginal distribution to infer the joint distribution would be a troublesome problem, since there are numerous joint distributions available to approach the targeted marginal distribution. Therefore, based on this preliminary, we have to add another assumption for coping with this ill-posed problem. The UNIT framework, which has proposed a term called ‘share-latent space’, indicating that it is possible to map a set of cross-domain images into the identical latent space.

\begin{figure}[h]
  \centering
    \includegraphics[width=0.5\textwidth]{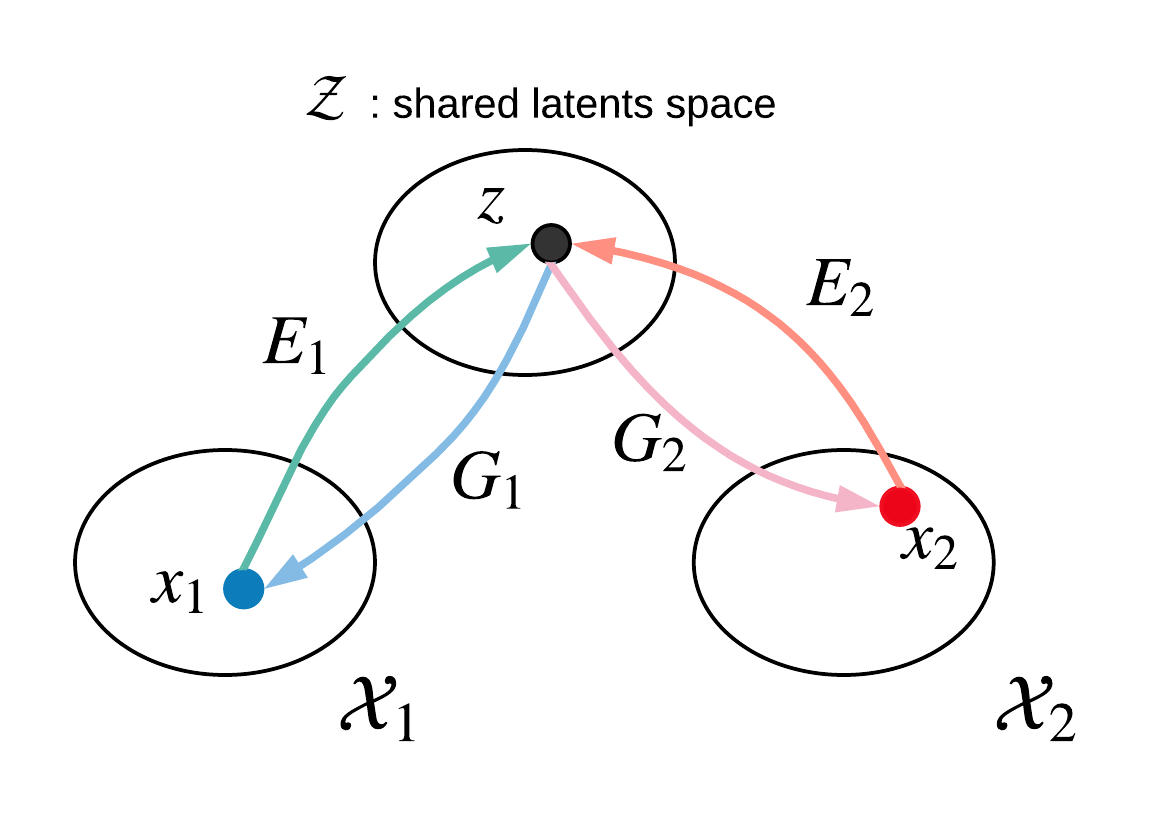}
    \caption[Shared Latent Space Assumption]{The shared latent space assumption~\citep{liu2017unsupervised}}
	\label{fig:assumption}
\end{figure}

As shown in Figure \ref{fig:assumption}, UNIT assumes that any pair of images $\{x_1,x_2\}$ coming from two different domains $\mathcal{X}_1$ and $\mathcal{X}_2$ can be mapped to the same latent code $z$ in the shared latent space $\mathcal{Z}$ by the encoding function $E_1$ and $E_2$ and mapped back by the decoding function $G_1$ and $G_2$. Postulated, there are four existing functions $E^{\ast}_1$, $E^{\ast}_2$, $G^{\ast}_1$, $G^{\ast}_2$ such that we have $z=E^{\ast}_1(x_1)=E^{\ast}_2(x_2)$ and conversely $x_1=G^{\ast}_1(z)$ and $x_2=G^{\ast}_2(z)$.  In this situation, the translation function can be seem as $F_{1\rightarrow 2}^{\ast}(x_1)=G_2^{\ast}(E_1^{\ast}(x_1))$ and $F_{2\rightarrow 1}^{\ast}(x_1)=G_1^{\ast}(E_2^{\ast}(x_2))$. 

However, directly generating the shared-latent code from two different domains will be difficult so the UNIT extends the previous shared-latent space assumption by adding an intermediate representation $h$~(i.e. Equation \ref{eq:inter}).
\begin{equation}
\label{eq:inter}
z\rightarrow h
  \begin{array}{l@{\ }l}
    \raisebox{-1ex}{$\nearrow$} &{x_1}\\
    \raisebox{1ex}{$\searrow$}  &{x_2}
   \end{array}
\end{equation}
With intermediate representation $h$, the Generators can be rewrite as $G_1^{\ast}\equiv G_{L,1}^{\ast}\circ G_{H}^{\ast}$ and $G_2^{\ast}\equiv G_{L,2}^{\ast}\circ G_{H}^{\ast}$ where $G_{H}$ represents the common high-level function that map the images from intermediate space $h$ to latent space $z$, and $G_{L,1}^{\ast},G_{L,2}^{\ast}$ represent the low-level generators that map images to the intermediate space $h$. Conversely, the encoder function can be rewritten as $E_1^{\ast}\equiv E_H^{\ast}\circ E_{L,1}^{\ast}$ and $E_2^{\ast}\equiv E_H^{\ast}\circ E_{L,2}^{\ast}$.  

In the real image translation tasks~(e.g. day and night image translation), $z$ can be regarded as the high-level representation of a scene~(e.g. ``bridge in back", ``bus in front"), and $h$ can be seem as the realisation of $z$ through $G_H^{\ast}$~(``bus occupy the following pixels"). Finally, the $G_{L,1}^{\ast},G_{L,2}^{\ast}$ will be the real image generation function~(e.g. ``bus is shining red during the day, but dark red in the night").

\section{Wasserstein Distance Guided Cross-Domain Learning}
\subsection{Problem in UNIT framework}
UNIT proposes an elegant Shared latent space hypothesis, but due to the VAE-based framework, it still has some potential problems such as latent code interaction and bad distance measurement. Recently, WAEs were proposed to solve the problem in VAEs-based framework by applying Wasserstein distance instead of KL-Divergence to measure discrepancy between different probability distributions which provides a much weaker topology and useful gradient information.

In this case, the same latent space assumption is followed; however the Wasserstein distance is applied to measure the discrepancy across different distributions. With the good gradient property and promising bound provided by Wasserstein distance, the framework ``Wasserstein Distance Guided Cross-Domain Learning~(WDGCDL)" constructs an elegant high quality latent space and achieves state-of-the-art results on domain adaptation benchmark datasets. 

\subsection{Net Architecture}	
Followed by the shared-latent space assumption, the model is proposed with a Wasserstein Auto-Encoder~(WAE) and Generative Adversarial Network~(GAN) based structure. Since the Wasserstein auto-encoder contains two type of constrain objectives, two different network architectures are proposed for different constrains.

\textbf{MMD-based Net:}
\begin{figure}[h] 
\centering
\includegraphics[width=0.7\textwidth]{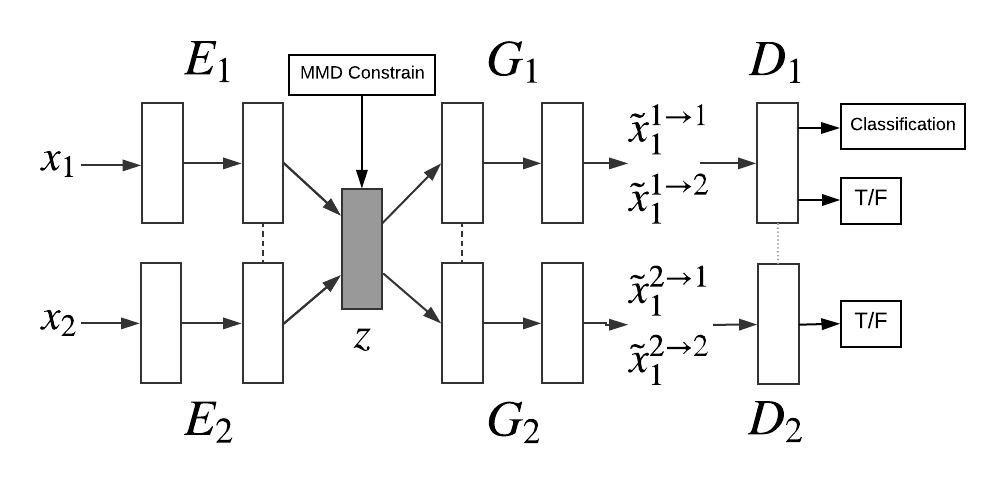}
\caption[WDGCDL Net Architecture~(MMD Constrain)]{WDGCDL Net Architecture~(MMD-based constrain): The solid line in the figure indicates the feed forward direction and the dashed line indicates the shared-weight layers}
\label{fig:wdgcdlm}
\end{figure}

Figure \ref{fig:wdgcdlm} illustrates the MMD-based Network architecture, the latent space will be constructed by minimising the maximum mean discrepancy~(MMD) loss between the generated latent code with prior latent code. 

\textbf{GAN-based Net:}
\begin{figure}[h] 
\centering
\includegraphics[width=0.7\textwidth]{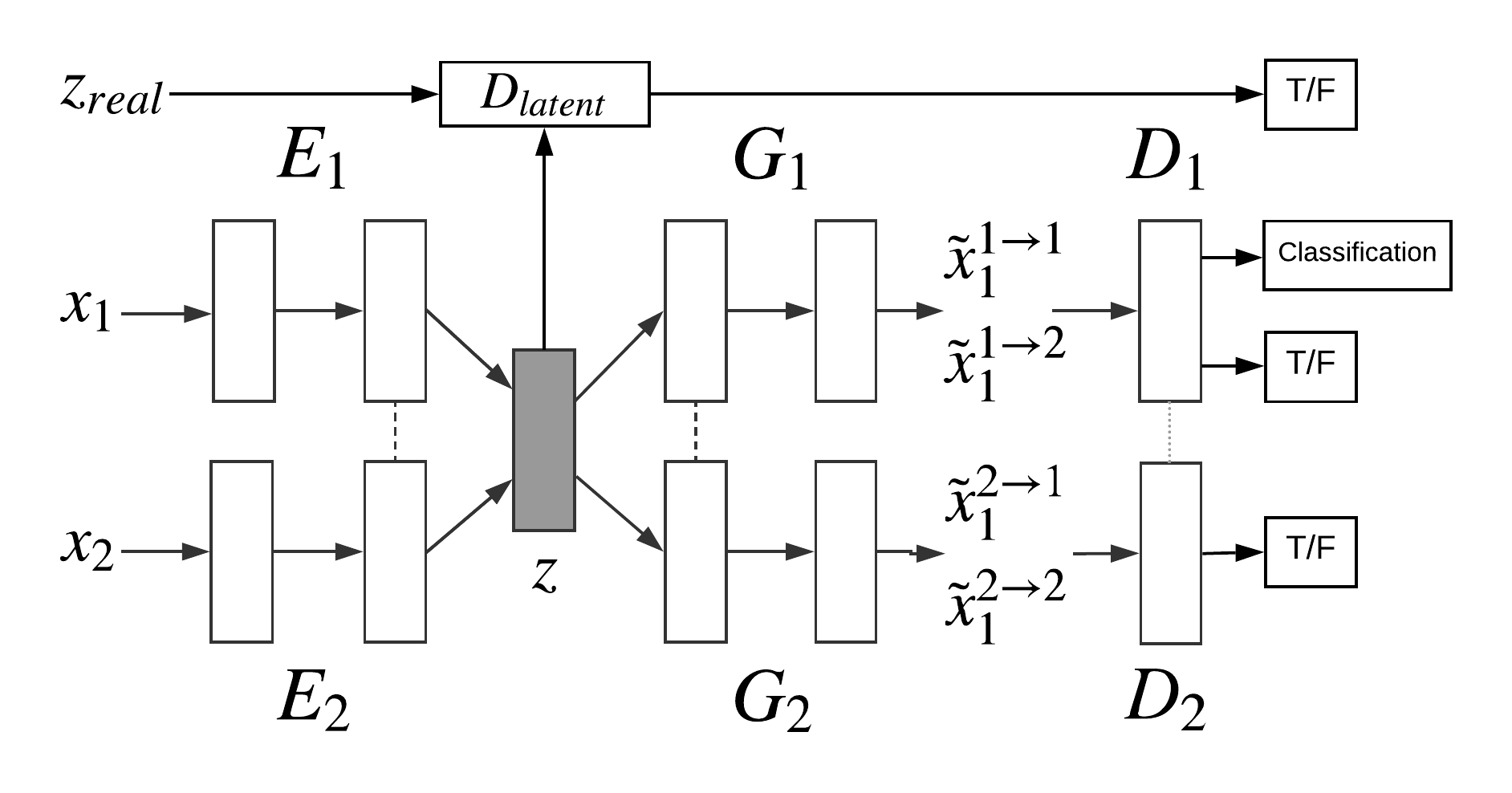}
\caption[WDGCDL Net Architecture~(GAN Constrain)]{WDGCDL Net Architecture~(GAN-based constrain)}
\label{fig:wdgcdlg}
\end{figure}

Figure \ref{fig:wdgcdlg} presents the adversary version of the framework. Instead of minimizing the MMD loss, the MMD constrain is replaced by attaching a latent discriminator $D_{latent}$ which aims to distinguish the fake latent code (i.e generated by encoders) from the real latent code (i.e. sample from Gaussian). In the final states, the generated latent code will be indistinguishable from the prior latent code.


To address the implementation of the intermediate representation $h$ and the shared-latent space assumption, the UNIT framework is followed and the weight-sharing constrain is enforced on the WAEs parts. The weight-sharing constrain is achieved by adding the shared-weight layer at the back of the encoder function and front layer of generator function.

Table \ref{tb:subnet} lists out the subnetworks in the our framework and their roles.  
\begin{table}[h]
\centering
\caption[Subnetworks Interpretation]{Interpretation of the roles of the subnetworks in the proposed framework}
\label{tb:subnet}
\begin{tabular}{@{}c|cccc@{}}
\toprule
Networks & $\{E_1,G_1\}$           & $\{G_1,D_1\}$           & $\{E_1,G_1,D_1\}$ & $\{G_1,G_2,D_1,D_2\}$ \\ \midrule
Roles    & WAE for $\mathcal{X}_1$ & GAN for $\mathcal{X}_1$ & VAE-GAN           & CoGAN                 \\ \bottomrule
\end{tabular}
\end{table}

The network details will be listed on Appendix \ref{appendix:a}.

\subsection{Objective Functions}
The objective of the framework will be divided into several parts based on the subnet structure. For the domain adaptation task, the framework will jointly solve the training problems of $WAE_1$, $WAE_2$, $GAN_1$, $GAN_2$ and the classification problem in $D_1$.

 \subsubsection{WAEs part Objective Functions} 
 WAEs training aims to minimise the Optimal Transport(OT) cost or Wasserstein distance between two distributions. In this case, the original objective in WAEs is followed and give out our objectives:
\begin{align}
\costfunc_{WAE_1}(E_1,G_1):=\lambda_0\cdot \underset{Q(z_1|x_1)\in P_{z_1}}{inf}\mathbb{E}_{P_{x_1}}\mathbb{E}_{Q(z_1|x_1)}[c(x_1,G_1(z_1))] + \lambda_1\cdot D_{Z}(Q_{Z_1},P_{Z_1})\\
\costfunc_{WAE_2}(E_2,G_2):= \lambda_0\cdot \underset{Q(z_2|x_2)\in P_{z_2}}{inf}\mathbb{E}_{P_{x_2}}\mathbb{E}_{Q(z_2|x_2)}[c(x_2,G_2(z_2))] + \lambda_1\cdot D_{Z}(Q_{Z_2},P_{Z_2})
\end{align}
where the $c(\cdot,\cdot)$ could be any distance measurement function. In the experiments, the $L_1$ distance is applied. The hyper-parameter $\lambda_0$ and $\lambda_1$ aims to control the weight of the objective terms where the regularisation on constrain term punish the deviation on the distribution of the latent code from prior distribution which makes the sample process more easier. The prior distribution is a zero mean Gaussian distribution where  $P_Z=\mathcal{N}(z|0,I)$. 

For the $D_Z$ constrain, the constrain on the WAEs framework is used: Maximum Mean Discrepancy based constrain~(Equation \ref{eq:mmd}) and GAN based constrain~(Equation \ref{eq:gan}). This constrains aims to force the encoding probability distribution to match the prior distribution via training process.

\textbf{MMD-based} $D_Z$: For a positive-definite reproducing kernel $k:\mathcal{Z}\times\mathcal{Z}\rightarrow \mathcal{R}$ the constrain can be written as:
\begin{equation}
\label{eq:mmd}
	MMD_k(P_Z,Q_Z)=||\int_\mathcal{Z}k(z,\cdot)dP_Z(z)-\int_\mathcal{Z}k(z,\cdot)dQ_Z(z)||_{\mathcal{H}_k}
\end{equation}

\textbf{GAN-based} $D_Z$: By introducing the adversarial procedure, the GAN-based constrain objective can be written as:
\begin{equation}
\label{eq:gan}
	GAN(P_Z,Q_Z)= \mathbb{E}_{z\sim P_{Z}}[log(z)]+\mathbb{E}_{\tilde{z}\sim Q_Z}[\log(1-\tilde{z})]
\end{equation}
where $z$ is the real latent code sample from Gaussian distribution and $\tilde{z}$ is the fake latent code generated by the encoding network.

 \subsubsection{GANs part Objective Functions} 
 GANs training aims to ensure that the translated images and reconstructed images are indistinguishable from the target domain and source domain respectively which can be formulated as:
 \begin{align}
  &\begin{aligned}
\costfunc_{GAN_1}(E_1,G_1,D_1) &= \lambda_2\mathbb{E}_{x_1\sim p_{\mathcal{X}_1}}[log D_1(x_1)]+\lambda_2\mathbb{E}_{z_1\sim q_1(z_1|x_1)}[\log(1-D_1(G_1(z_1)))]\\
	&\quad +\lambda_2\E_{z_2\sim q_2(z_2|x_2)}[\log(1-D_1(G_1(z_2)))]
  \end{aligned}\\
  &\begin{aligned}
\costfunc_{GAN_2}(E_2,G_2,D_2) &= \lambda_2\mathbb{E}_{x_2\sim p_{\mathcal{X}_2}}[log D_2(x_2)]+\lambda_2\mathbb{E}_{z_2\sim q_2(z_2|x_2)}[\log(1-D_2(G_2(z_2)))]\\&\quad+\lambda_2\E_{z_1\sim q_1(z_1|x_1)}[\log(1-D_2(G_2(z_1)))]
  \end{aligned}
  \end{align}
  The hyper-parameter $\lambda_2$ aims to control the impact of the GAN object.
\subsubsection{Adaptation Objective Functions} 
To achieve the domain adaptation task, the weight of $D_1$ and $D_2$ were tied. This allowed me to adapt a classifier trained in the source domain to the target domain. Moreover, I enforced the $L_1$ distance between the features extracted by the front layers of discriminator between $D_1$ and $D_2$ which encouraged these two discriminators to interpret the images from different domains in the same way. Defining the feature maps extracted by the front layers of Discriminators as $\{h^s,h^t\} =\{D^{front}_{s}(x_s),D^{front}_{t}(x_t)\}$ and $\{h^{s\rightarrow t},h^{t\rightarrow s}\} =\{D^{front}_{s}(x_s^{s\rightarrow t}),D^{front}_{t}(x_t^{t\rightarrow s})\}$ . The constrain can be formulated as:
\begin{align}
	\costfunc_{D_{F}}(D_1) &=\lambda_3 \cdot \mathbb{E}_{x_s\sim p_{data}(X^{\mathcal{S}})}[||h^s-h^{s\rightarrow t}||_1]\\
\costfunc_{D_{F}}(D_2) &=\lambda_3 \cdot \mathbb{E}_{x_t\sim p_{data}(X^{\mathcal{T}})}[||h^t-h^{t\rightarrow s}||_1]
\end{align} 
The hyper-parameter $\lambda_3$ control the weight of the feature map objective term.

Finally, the shared classifier which constructed by the shared weight front layers in $D_1$ and $D_2$ and an unique classifier trained by the label from source domain. The objective can be formulated as a standard softmax cross-entropy form:
\begin{equation}
	\costfunc_{Cla}(C_1)= \lambda_4 \cdot\mathbb{E}_{\{x_1,y_1\}\sim \mathcal{S}}\left [ -\mathbf{y_1}\cdot log(\hat{\mathbf{y}}_1) \right ]
\end{equation}
where $\hat{\mathbf{y}}_1=C_1(G_{1}(E_{1}(x_1)))$ and $\mathbf{y}_{1}$ is the one-hot encoding of the class label $y_{1}$. The hyper-parameter $\lambda_4$ control the weight of the classification objective term.

In this case, the whole objectives of the framework can be formulated as:
\begin{align}
&\begin{aligned}
\underset{E_1,E_2,G_1,G_2,C_1}{min}~\underset{D_1,D_2}{max}~&\costfunc_{WAE_1}(E_1,G_1)+\costfunc_{GAN_1}(E_1,G_1,D_1)+\costfunc_{D_{F}}(D_1)+ \costfunc_{Cla}(C_1)\\
&\costfunc_{WAE_2}(E_2,G_2)+\costfunc_{GAN_2}(E_2,G_2,D_2)+\costfunc_{D_{F}}(D_2)
\end{aligned}
\end{align}

\subsection{Training Scheme}
Different from the UNIT framework, the applied Wasserstein distance in the latent space increase the complexity of the training process. In this case, I carefully design two training schemes for the proposed framework which can ensure the net can be trained in a stable way.  
\subsubsection{MMD-Based Training Algorithm}
The first one training scheme is the MMD-based objective training scheme. For the MMD kernel, I choose the inverse multi quadratic~(IMQ) kernel~(i.e. Equation \ref{eq:imq}).
\begin{equation}
\label{eq:imq}
	k(x,y) = \frac{1}{\sqrt{||x-y||^2-c^2}}
\end{equation}
During this training scheme, I first applied gradient descent step to update the discriminator $D_1$ and $D_2$ with $E_1,E_2,G_1$ and $G_2$ fixed. Then, I applied the gradient descent step to update the generator part~(i.e. $E_1,E_2,G_1$ and $G_2$) by fixing the discriminator parameters $D_1$ and $D_2$. More details of this algorithm will be listed in the Algorithm \ref{al:Wdgcdlmmd}.

\begin{algorithm}
\caption{Wasserstein Distance Guided Cross Domain Learning with MMD-based penalty(WDGCDL-MMD).}
\label{al:Wdgcdlmmd}
\begin{algorithmic}[lines]
\Require Regularization coefficient $\alpha,\beta,\gamma,\lambda,\sigma > 0$,  \\
characteristic positive-definite kernel $k$, \\
Initialize the parameters of the Generator $G_{\theta}$,\\
Initialize the parameters of the Discriminator $D_{\phi}$
   \While{$(\theta,\phi)$ not converge}
      \State Sample $\{(x^{s}_{i},x^{t}_{i})\}$, $\{y^s_i\}$ from the training set
      \State Sample shared $\{\tilde{z}_i\}$ from $G_{Encoder}(Z|(x^{s}_{i},x^{t}_{i}))$ for $i=1,...,n$
      \State Sample $\{\tilde{x}^{s}_{i}$,$\tilde{x}^{t}_{i}\}$,$\{\tilde{x}^{s\rightarrow t}_{i}$,$\tilde{x}^{t\rightarrow s}_{i}\}$ from $G_{\theta}(x^{s}_{i},x^{t}_{i})$
      \State Sample $\{h^{s}_i, h^{t}_i \}$,$\{\tilde{h}^{s}_i$,$\tilde{h}^{t}_i \}$,$\{\tilde{h}^{s\rightarrow t}_i, \tilde{h}^{t\rightarrow s}_i \}$from $D_{\phi}(x^{s}_{i},x^{t}_{i})$, $D_{\phi}(\tilde{x}^{s}_{i},\tilde{x}^{t}_{i})$, $D_{\phi}(\tilde{x}^{s\rightarrow t}_{i},\tilde{x}^{t\rightarrow s}_{i})$
      \State Sample Sample classification output $\{\tilde{y}_{i}^{s} \}$ from $D_{\phi}(x_{i}^{s})$
      \State Sample shared latent variable $\{z_i\}$ from prior $\mathcal{P}_{Z}$
      \State Update $D_{\phi}$ by descending:
      \begin{align*}
      \begin{gathered}
      	\frac{\alpha}{n}\sum_{i}[\log D_{\phi}(x^s_i)+\log D_{\phi}(x^t_i)+\log(1-D_{\phi}(\tilde{x}^s_i))+\log(1-D_{\phi}(\tilde{x}^t_i))\\+\log(1-D_{\phi}(\tilde{x}^{s\rightarrow t}_i))+\log(1-D_{\phi}(\tilde{x}^{t\rightarrow s}_i))]+ \frac{\beta}{n}\sum_{i}[c(h^s_i, h^{s\rightarrow t}_i)+c(h^t_i, h^{t\rightarrow s}_i)] \\+\frac{\gamma}{n}\sum_{i}y_{i}^{s}\log(\tilde{y}_{i}^{s})
      \end{gathered}	 
      \end{align*}
	  
	  \State Update $G_{\theta}$ by descending:\\
	  \begin{align*}
      \begin{gathered}
	  \frac{\alpha}{n}\sum_{i}[\log D_{\phi}(\tilde{x}^s_i)+\log D_{\phi}(\tilde{x}^t_i)+\log D_{\phi}(\tilde{x}^{s\rightarrow t}_i)+\log D_{\phi}(\tilde{x}^{t\rightarrow s}_i)]\\
	  +\frac{\lambda}{n}\sum_i[c(x^s_i,\tilde{x}^s_i)+c(x^t_i, \tilde{x}^t_i)]+ \frac{\sigma}{n(n-1)}\sum_{i\neq j}k(z_{i},z_j)+\frac{\sigma}{n(n-1)}\sum_{i\neq j}k(\tilde{z}_i,\tilde{z}_j)\\+\frac{2\sigma}{n^2}\sum_{i\neq j}k(z_i,\tilde{z}_j)
	  \end{gathered}	 
      \end{align*}
   \EndWhile
\end{algorithmic}
\end{algorithm}

\subsubsection{GAN-Based Training Algorithm}
For the GAN-Based training scheme, I added another discriminator in the latent space so that the training process will become more complex. I first fixed the parameters of $E_1,E_2,G_1,G_2$ and $D_{latent}$ and applied gradient descent step to update the discriminator $D_1$ and $D_2$. Then, I apply gradient descent step to update the latent discriminator $D_{latent}$ with fixed parameters in $E_1,E_2,G_1,G_2,D_1$ and $D_2$. Finally, I fixed the parameters in all the discriminator parts and applied gradient descent step to the generator parts $E_1,E_2,G_1$ and $G_1$. More details of this algorithm will be listed in Algorithm \ref{al:Wdgcdlgan}.

\begin{algorithm}
\caption{Wasserstein Distance Guided Cross Domain Learning with GAN-based penalty(WDGCDL-GAN).}
\label{al:Wdgcdlgan}
\begin{algorithmic}[lines]
\Require Regularization coefficient $\alpha,\beta,\gamma,\lambda,\eta> 0$,  \\
Initialize the parameters of the Generator $G_{\theta}$,\\
Initialize the parameters of the Discriminator $D_{\phi}$,\\
Initialize the parameters of the Latent Discriminator $D_{\tau}$,
   \While{$(\theta,\phi,\tau)$ not converge}
      \State Sample $\{(x^{s}_{i},x^{t}_{i})\}$, $\{y^s_i\}$ from the training set
      \State Sample shared $\{\tilde{z}_i\}$ from $G_{Encoder}(Z|(x^{s}_{i},x^{t}_{i}))$ for $i=1,...,n$
      \State Sample $\{\tilde{x}^{s}_{i},\tilde{x}^{t}_{i}\},\{\tilde{x}^{s\rightarrow t}_{i},\tilde{x}^{t\rightarrow s}_{i}\}$ from $G_{\theta}(x^{s}_{i},x^{t}_{i})$
      \State Sample $\{h^{s}_i, h^{t}_i \}$,$\{\tilde{h}^{s}_i$,$\tilde{h}^{t}_i \}$,$\{\tilde{h}^{s\rightarrow t}_i, \tilde{h}^{t\rightarrow s}_i \}$from $D_{\phi}(x^{s}_{i},x^{t}_{i})$, $D_{\phi}(\tilde{x}^{s}_{i},\tilde{x}^{t}_{i})$, $D_{\phi}(\tilde{x}^{s\rightarrow t}_{i},\tilde{x}^{t\rightarrow s}_{i})$
      \State Sample Sample classification output $\{\tilde{y}_{i}^{s} \}$ from $D_{\phi}(x_{i}^{s})$
      \State Sample shared latent variable $\{z_i\}$ from prior $\mathcal{P}_{Z}$
      \State Update $D_{\phi}$ by descending:
      \begin{align*}
      \begin{gathered}
      	\frac{\alpha}{n}\sum_{i}[\log D_{\phi}(x^s_i)+\log D_{\phi}(x^t_i)+\log(1-D_{\phi}(\tilde{x}^s_i))+\log(1-D_{\phi}(\tilde{x}^t_i))\\+\log(1-D_{\phi}(\tilde{x}^{s\rightarrow t}_i))+\log(1-D_{\phi}(\tilde{x}^{t\rightarrow s}_i))]+ \frac{\beta}{n}\sum_{i}[c(h^s_i, h^{s\rightarrow t}_i)+c(h^t_i, h^{t\rightarrow s}_i)] \\+\frac{\gamma}{n}\sum_{i}y_{i}^{s}\log(\tilde{y}_{i}^{s})
      \end{gathered}	 
      \end{align*}
      \State Update $D_{\tau}$ by descending:
      \begin{align*}
      	\begin{gathered}
      	\frac{\eta}{n}\sum_{i}[\log D_{\tau}(z_i)+\log(1-D_{\tau}(\tilde{z}_i))]
      	\end{gathered}	 
      \end{align*}
      \State Update $G_{\theta}$ by descending:
      \begin{align*}
      \begin{gathered}
	  \frac{\alpha}{n}\sum_{i}[\log D_{\phi}(\tilde{x}^s_i)+\log D_{\phi}(\tilde{x}^t_i)+\log D_{\phi}(\tilde{x}^{s\rightarrow t}_i)+\log D_{\phi}(\tilde{x}^{t\rightarrow s}_i)]\\
	  +\frac{\lambda}{n}\sum_i[c(x^s_i,\tilde{x}^s_i)+c(x^t_i, \tilde{x}^t_i)]+ \frac{\eta}{n}\log D_{\tau}(\tilde{z}_i)
	  \end{gathered}	 
      \end{align*}
   \EndWhile
\end{algorithmic}
\end{algorithm}

\chapter{Evaluation}

\section{Benchmark Dataset}

\subsection{Street View House Numbers(SVHN)}
\begin{figure}[h]
  \centering
\subfigure[SVHN Dataset Samples]{\label{fig:svhn}\includegraphics[width=0.315\textwidth]{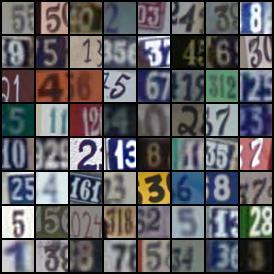}}
\subfigure[MNIST Dataset Samples]{\label{fig:mnist}\includegraphics[width=0.315\textwidth]{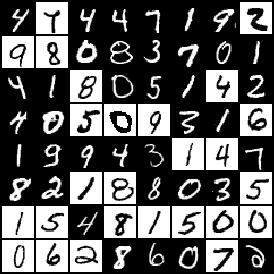}}
\subfigure[USPS Dataset Samples]{\label{fig:usps}\includegraphics[width=0.315\textwidth]{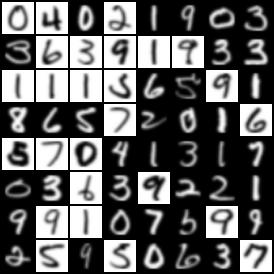}}
\caption[SVHN,MNIST,USPS Dataset Samples]{SVHN,MNIST,USPS Dataset Samples}
\label{fig:dz_image}
\end{figure}

The Street View House Numbers~(SVHN)~\citep{netzer2011reading} dataset is a MNIST-like dataset which contains images centred around a single character~(i.e. Figure \ref{fig:svhn}). In these experiments, I used the extra training dataset in SVHN dataset which contain 531131 images. 

\subsection{MNIST}
The MNIST handwritten digits dataset~\citep{lecun1998gradient} has a training set of 60000 examples, and a test set of 10000 examples. Figure \ref{fig:mnist} illustrates the image samples of MNIST dataset with the ground truth label from 0-9. As the MNIST images were in grey-scale, so that I convert it into RGB images and perform data argumentation on it~(i.e. inversions of original training dataset). The MNIST images size is 28$\times$28 pixels so that I resize MNIST dataset to 32$\times$32 for facilitating the experiments.

\subsection{USPS}
The USPS handwritten digits dataset~\citep{hull1994database} is similar to the MNIST dataset but have different shapes and more blurry. Figure \ref{fig:usps} shows the sample images from the USPS dataset. The USPS dataset is grey-scale, so that I convert it to RGB images and do the same argumentation as MNIST dataset. For experiments convenience, I also resize the USPS to 32$\times$32 size.

Moreover, I found that spatial context information was useful, so that I generate another two channels for each images which contain the normalised $x$ and $y$ coordinates.

\section{Network Initialisation}
The essence of training deep learning model is to update the weight~(i.e. parameter $\omega$), and that requires a corresponding initial value for each parameter. In the traditional machine learning problem or some convex function, we always set it to 0 and might perform well. However, for deep learning, nonlinear functions are madly superimposed which result in a non-convex function. Moreover, the starting point of initialisation will bring the following effects: a) Determine whether the algorithm will converge; b) Determine how fast learning can converge; c) Affect generalisation. In this case, the way to select the initial value of the parameters becomes a problem worthy of discussion. There are some state-of-the-art ways to do the weight initialization such as He initialization~\citep{he2015delving} and Xavier~\citep{glorot2010understanding} initialization.

In this framework, Gaussian initialisation~(i.e. randomly initialise weight) was applied for generator and Xavier initialisation for Discriminator part.
 
\section{Hyper-parameter Tuning}
In the experiment, there are various hyper-parameters that influence the performance of whole framework. In this case, to find out the best combination of hyper-parameters, I tried various different hyper-parameter combinations via greedy search. 

The framework consists of 7 hyper-parameters: latent space dimension~($d_z$), latent space distribution variance~($\sigma$), weight on Wasserstein distance or reconstruction loss~($\lambda_0$), weight on Constrain term~($\lambda_1$)(The experiments are mainly using MMD constrain, so that we will use MMD term instead of constrain term in the following paper), weight on GAN objective term~($\lambda_2$), weight on feature loss term~($\lambda_3$) and weight on classification loss term~($\lambda_4$). The following tuning process is based on MNIST $\rightarrow$ USPS experiment under MMD-based training scheme; all parameters setting are given at the end.
\subsection{Generator Hyper-parameter Tuning}
Because of the difficulty of adversarial training, I first tuned the hyper-parameter in generator without the discriminator in order to find out the suitable range of the hyper-parameters in the generator. There are two hyper-parameters, which mainly affect the output of generator: latent space dimension~($d_z$) and MMD constrain weight~($\lambda_1$). In this case, other hyper-parameters were fixed in generator with a suitable value and the latent space was tuned with MMD constrain via greedy search.

\textbf{Latent space dimension($d_z$)}: For the latent space dimension $d_z$, I follow the experiments in WAEs and set the range from 2 to 256 (i.e. $\{2,4,8,16,32,64,128,256\}$)

\textbf{MMD Constrain weight($\lambda_1$)}: For the MMD objective term, I carefully set it from 0.001 to 100 (i.e. $\{,0.01,0.01,0.1,1,10,100\}$)

As I set the range of hyper-parameters, I do the greedy search for latent space dimension and the MMD constrain with other hyper-parameters fixed with a suitable values(e.g. $\lambda_0=0.1,\sigma=2$). For the greedy search, I first fix the latent space dimension and try different MMD constrain until find out the best one. 

\begin{figure}[h]
	\centering
		\includegraphics[width=1\textwidth]{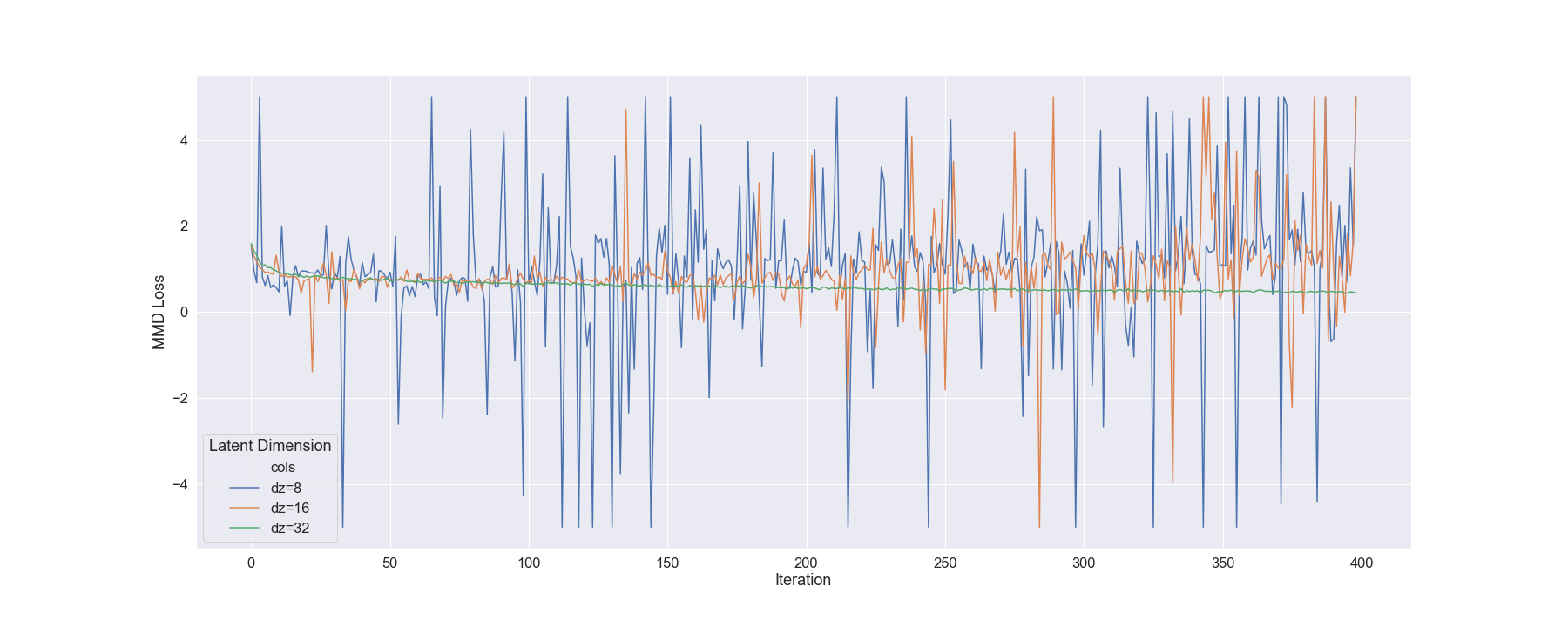}
	\caption[MMD Loss change under different Latent space Dimension]{Different latent space dimensions' impact on MMD loss descent}
	\label{fig:mmd_ls}
\end{figure}

During the tuning process, I found that smaller latent space dimension~(e.g. $d_z=2,4,8$) will led the divergence or fluctuation of MMD constrain loss for any constrain weight. The potential reason is that low dimensional latent space does not have enough spaces to learn the information. In this case, I fixed the MMD weight and tried higher latent dimension $d_z$. Figure \ref{fig:mmd_ls} illustrates the MMD loss change and reconstruction result after 400 iteration training processes.

\begin{figure}[h]
\centering     
\subfigure[Latent Dimension $d_z=8$]{\label{fig:8dz}\includegraphics[width=0.315\textwidth]{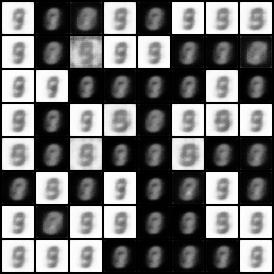}}
\subfigure[Latent Dimension $d_z=16$]{\label{fig:16dz}\includegraphics[width=0.315\textwidth]{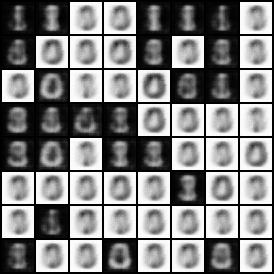}}
\subfigure[Latent Dimension $d_z=32$]{\label{fig:32dz}\includegraphics[width=0.315\textwidth]{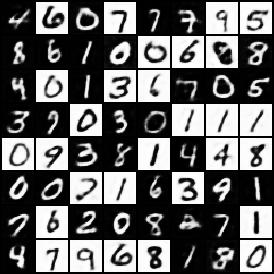}}
\caption[Reconstruction images quality with different latent space dimension]{Reconstruction images quality with different latent space dimension}
\label{fig:dz_image}
\end{figure}

From \fref{fig:mmd_ls}, it is found that the descent of MMD loss become more stable~(i.e. Green line) with the increasing of latent space dimension. Moreover, the reconstruction digital numbers shape become more complex from $d_z=8$ to $d_z=32$(i.e. \fref{fig:dz_image}).

Using this method, I also found out the potential latent space for SVHN $\rightarrow$ MNIST experiment which is $d_z=64$. However, the adversarial training will influence the loss so that there is still need to find the best latent space dimension after adding the discriminator. Fortunately, after the generator parameter search, the direction of tuning the latent space dimension can be known.

\subsection{Complete Net Hyper-parameters Tuning}
After finding out the potential latent space dimension in generator, I add the discriminator for the adversarial training and start the hyper-parameter search for entire net.

\textbf{Tuning Difficulty:} The hyper-parameter search under adversarial training will become difficult as the signal from discriminator will influence the generator, and vice versa. Moreover, if the generator becomes too perfect~(i.e. easy to generate real-like images) then the discriminator will receive useless information for training. In contrast, if the discriminator becomes too perfect~(i.e. easy to distinguish the fake images and real images), then the generator can not receive useful gradient for training process. In this case, I aimed to find the trade-off situation where both generator and discriminator were not too perfect.

\textbf{Latent Space Dimension($d_z$):} As we wanna the discriminator can receive the useful signal from the generator so that we need to make sure our generator can lead the correct training direction for discriminator. In this case, I followed the tuning experiment on generator and first tuning the latent dimension space $d_z$. From the previous experiments, I narrowed the tuning interval of latent space dimension from $\{2,4,8,16,32,64,128,256\}$ to $\{32,64,128,256\}$. To tuning the latent dimension space, we need to fix other hyper-parameters as other parameters will affect the performance. In this case, I followed the parameter setting in the UNIT framework for the discriminator part. The fixed hyper-parameters: GAN objective($\lambda_2=10$), Feature loss weight($\lambda_3=0.0001$), Classification loss weight($\lambda_4=10$), Reconstruction loss weight($\lambda_0=0.01$) and MMD loss weight($\lambda_1=1$). I first tried the potential latent space dimension from generator tuning process~(i.e. $d_z=32$) but the reconstruction image was irregular. In this case, I increased the latent space from $d_z=32$ to 64, which gave a good reconstructed image. I applied the same operation to the SVHN $\rightarrow$ MNIST experiment~(i.e.  increase $d_z$ from 64 to 128). 

The Latent space dimension hyper-parameter are the basis of the parameter search, all of other parameters tuning will based on the true latent space dimension. For the rest of the parameters, we are still going to do greedy search but we have got some basic directions. For example, the classification loss weight, feature loss weight, reconstruction loss weight and MMD loss weight are in a stable range, so we will put the adjustment to them at the end. Different with the previous tuning step, the evaluation way of rest parameters will change to the translated image quality and classification accuracy on target domain.

\textbf{GAN Loss Weight($\lambda_2$):} As I intended to get a perfect classifier under the adversarial training process, the GAN weight loss is more important as it controls the main direction of training process. In this case, I first tuned the GAN loss weight by fixing other parameters. I set the tuning interval from 100 to 0.1 (i.e. $\{100,10,1,0.1\}$)and ran the greedy search.

\begin{figure}[h]
	\centering
		\includegraphics[width=0.8\textwidth]{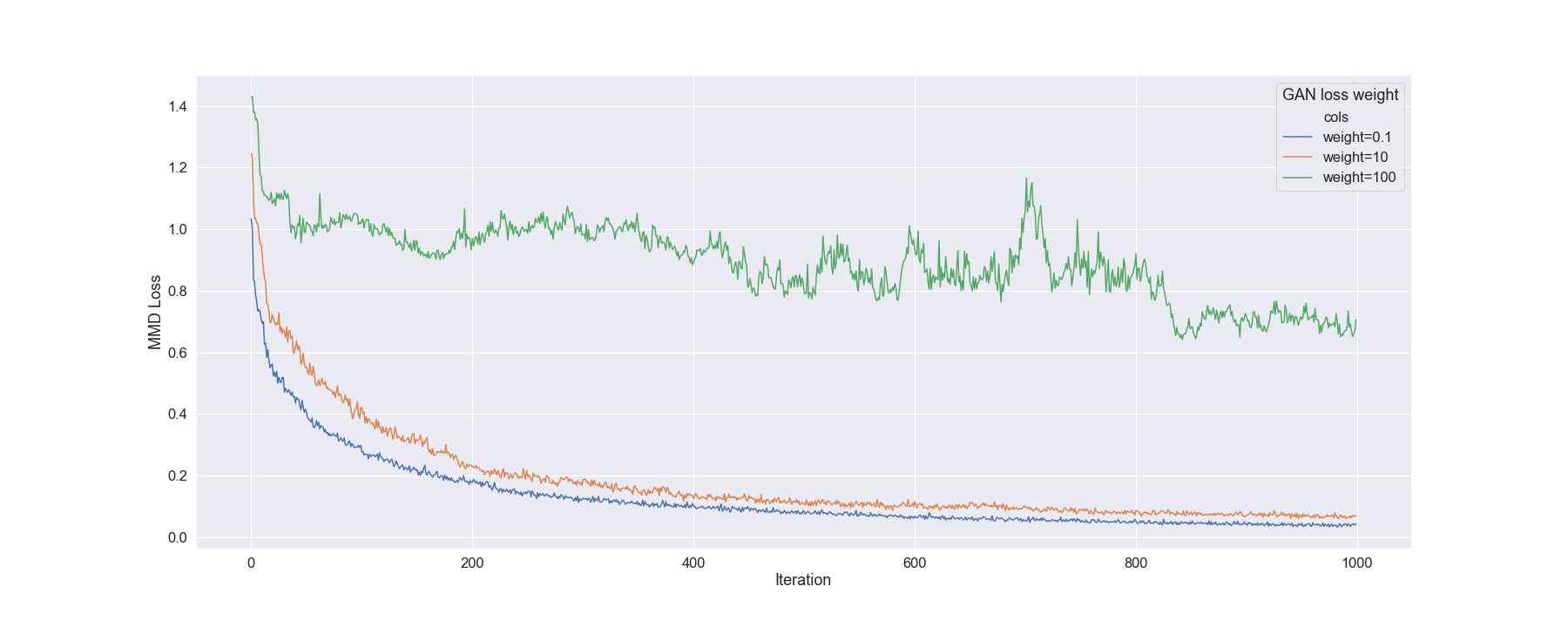}
	\caption[MMD loss descent situation under different GAN loss weights]{MMD loss descent situation under different GAN loss weights}
	\label{fig:mmd_ganw}
\end{figure}

\begin{figure}[h]
	\centering
		\includegraphics[width=0.9\textwidth]{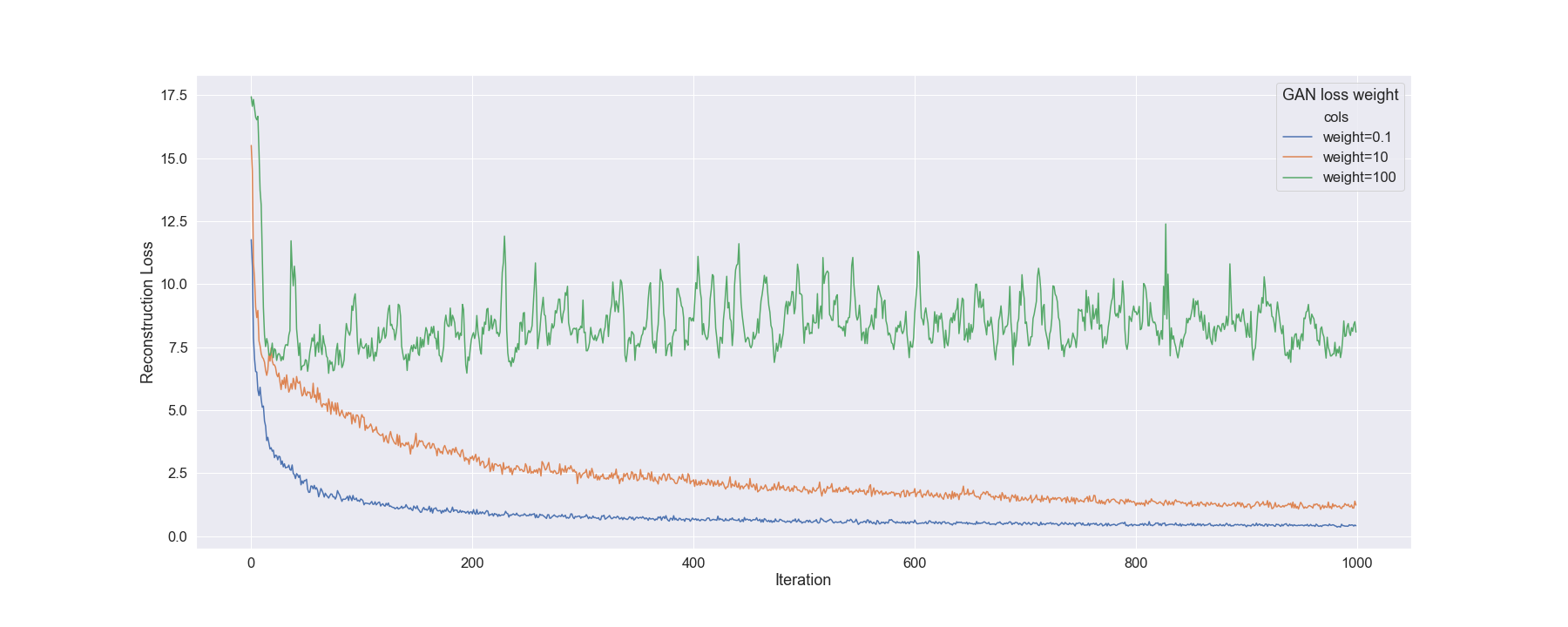}
	\caption[Reconstruction loss descent situation under different GAN loss weights]{Reconstruction loss descent situation under different GAN loss weights}
	\label{fig:recon_ganw}
\end{figure}

During the tuning process, I found that the convergence speed of other key loss~(i.e. MMD loss, Reconstruction loss) was faster and more stable as the GAN loss weight is reduced. Figure \ref{fig:mmd_ganw} and \ref{fig:recon_ganw} illustrate the MMD loss and reconstruction loss change after 1000 iteration training process.

\begin{figure}[h]
	\centering
		\includegraphics[width=0.9\textwidth]{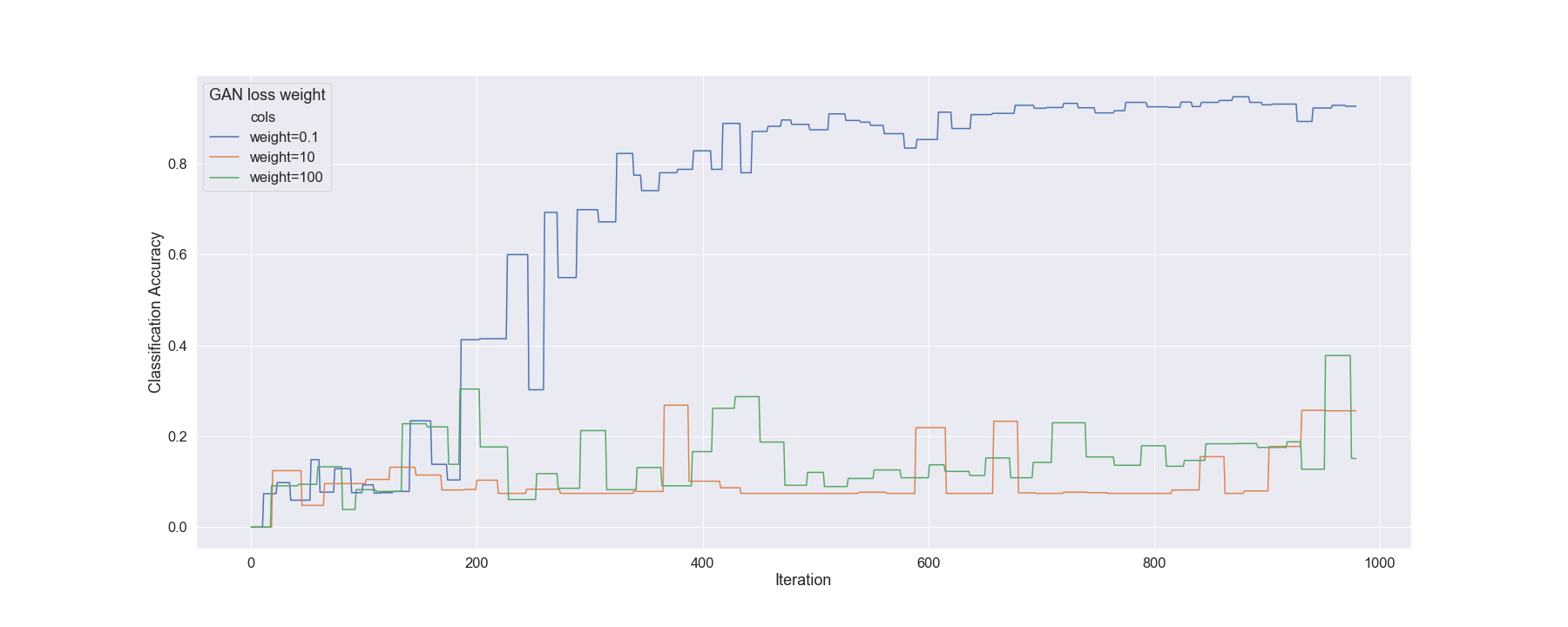}
	\caption[Classification Accuracy under different GAN loss weights]{Classification Accuracy under different GAN loss weights}
	\label{fig:cla_ganw}
\end{figure}

Moreover, the accuracy of classification will be greatly improved when $\lambda_2=0.1$(i.e. \fref{fig:cla_ganw}). It is possible that the reason for this situation is that the larger loss weight results in a larger value when calculating the gradient, thus failing to reach the global or local optimal point.\\

\textbf{Other Loss weight:} For the rest parameters, I fine-tuned them until I got the best results. Table \ref{tb:parameter} will lists the best hyper-parameters for each experiments.

\begin{table}[h]
\centering
\begin{tabular}{@{}cccc@{}}
\toprule
Hyper-parameters                       & SVHN$\rightarrow$MNIST & MNIST$\rightarrow$USPS & USPS$\rightarrow$MNIST \\ \midrule
Latent Dimension $d_z$                 & 128                    & 64                     & 64                     \\
Latent Variance $\sigma$               & 2                      & 2                      & 2                      \\
Reconstruction Loss weight $\lambda_0$ & 0.2                    & 0.01                   & 0.01                   \\
MMD Loss weight $\lambda_1$            & 1                      & 0.2                    & 0.2                    \\
GAN Loss weight $\lambda_2$            & 20                     & 0.1                    & 0.1                    \\
Feature Constrain weight $\lambda_3$   & 0.001                  & 0.001                  & 0.001                  \\
Classification Loss weight $\lambda_4$ & 10                     & 10                     & 10                     \\ \bottomrule
\end{tabular}
\caption[Hyper-parameter setting for all expriments]{Hyper-parameter setting for all expriments}
\label{tb:parameter}
\end{table}
\subsection{Training Parameter Setting}
Apart from the network configuration, I also tuned to training parameters to ensure that the net could be trained in a stable way. In this case, I set learning rate to a small value $lr=0.0001$ and apply the state-of-the-art Adaptive Moment Estimation~(Adam)~\citep{kingma2014adam} optimiser. Moreover, I set decay $weight\ decay=0.0005$ to the optimiser, as it can punish the weight of useless item to prevent the overfitting problem. For the batch size of dataset, I set training batch size as 64 and testing batch size as 100. The maximum iteration of training process was 200000, I tested the classification performance on test dataset after every 100 iterations to ensure that the training process was on the right direction.

\section{Result}
All the experiments were run on the Google cloud virtual machine. The configuration of virtual machine is 8 vCPU with 32 Gigabit memory, 1 NVIDIA Tesla K80 GPU with 12 Gigabit memory, Ubuntu 16.04 LTS system.

Table \ref{tb:result} shows the comparison between the proposed model and some state-of-the-art models. With the elegant shared-latent space constructed by the Wasserstein distance, our proposed framework outperform the state-of-the-art models~\citep{liu2017unsupervised} in MNIST $\leftrightarrow$ USPS domain adaptation tasks. In the SVHN $\rightarrow$ MNIST experiment, the reason that it didn't outperform the other models is the time limitation of finding best hyper-parameters combination for this tasks. 
\begin{table}[h]
\centering
\label{tb:result}
\begin{tabular}{@{}cccc@{}}
\toprule
METHOD      & SVHN $\rightarrow$ MNIST & MNIST $\rightarrow$ USPS & USPS $\rightarrow$ MNIST \\ \midrule
Source Only & 0.6010                   & 0.7520                   & 0.5710                   \\
SA          & 0.5932                   & -                        & -                        \\
DANN        & 0.7385                   & -                        & -                        \\
DTN         & 0.8488                   & -                        & -                        \\
ADDA        & 0.7600                   & 0.8940                   & 0.9010                   \\
CoGAN       & -                        & 0.9565                   & 0.9315                   \\
UNIT        & 0.9053                   & 0.9597                   & 0.9358                   \\
WDGCDL-MMD(proposed) & 0.7847                   & \textbf{0.9731}          & \textbf{0.9372}          \\ \midrule
Target Only & 0.9630                   & 0.9920                   & 0.9920                   \\ \bottomrule
\end{tabular}
\caption[Unsupervised domain adaptation performance]{Unsupervised domain adaptation performance. The reported numbers are classification accuracies.}
\end{table}

From the result, it is found that the Wasserstein distance truly provides much more useful gradient information than the KL-divergence when measuring the discrepancy of distribution.

As the domain adaptation work were supported by the image-to-image translation task. Instead of the nice results on adaptation work, I also get some results on the image-to-image translation task.

\subsection{SVHN$\rightarrow$MNIST Translated Image Result}

\begin{figure}[h]
\centering   
\subfigure[Real Images]{\includegraphics[width=0.3\textwidth]{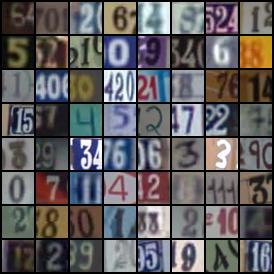}}
\subfigure[Reconstructed Images]{\includegraphics[width=0.3\textwidth]{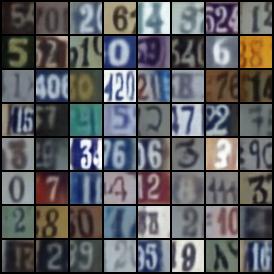}}
\subfigure[Translated Images]{\includegraphics[width=0.3\textwidth]{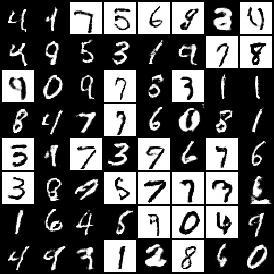}}
\caption[SVHN$\rightarrow$MNIST Result(On SVHN Domain)]{SVHN$\rightarrow$MNIST Result(On SVHN Domain)}
\label{fig:s2ms}
\end{figure}

\begin{figure}[h]
\centering
\subfigure[Real Images]{\includegraphics[width=0.3\textwidth]{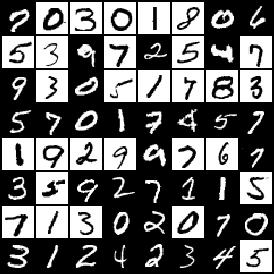}}
\subfigure[Reconstructed Images]{\includegraphics[width=0.3\textwidth]{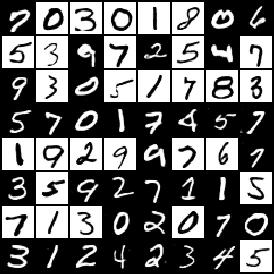}}
\subfigure[Translated Images]{\includegraphics[width=0.3\textwidth]{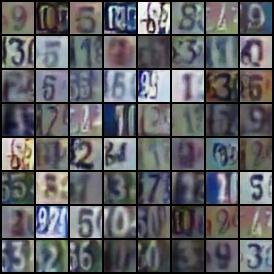}}
\caption[SVHN$\rightarrow$MNIST Result(On MNIST Domain)]{SVHN$\rightarrow$MNIST Result(On MNIST Domain)}
\label{fig:s2mm}
\end{figure}

For SVHN$\rightarrow$MNIST experiment, since we did not find the optimal parameters, the translation result was not so perfect. Additionally, \fref{fig:s2ms} and \ref{fig:s2mm} introduce the reconstructed and translated images result of this experiment. The images from left are ``Real Image", ``Reconstructed Image" and ``Translated Image".

\subsection{MNIST$\leftrightarrow$USPS Translated Image Result}
For the MNIST$\leftrightarrow$USPS experiment, since we find the best parameter setting and outperform the state-of-the-art models in domain adaptation work. The quality of translated image will higher than the SVHN $\rightarrow$ MNIST experiment. The following figures illustrate the results of experiments.

\textbf{MNIST$\rightarrow$USPS:}

\begin{figure}[h]
\centering   
\subfigure[Real Images]{\includegraphics[width=0.3\textwidth]{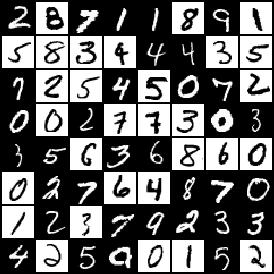}}
\subfigure[Reconstructed Images]{\includegraphics[width=0.3\textwidth]{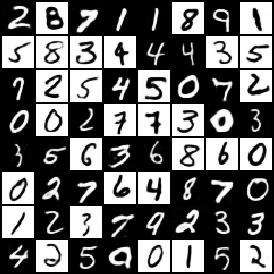}}
\subfigure[Translated Images]{\includegraphics[width=0.3\textwidth]{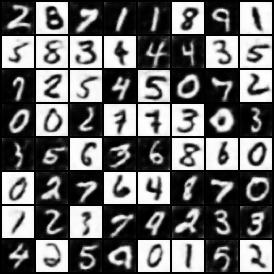}}
\caption[MNIST$\rightarrow$USPS Result(On MNIST Domain)]{MNIST$\rightarrow$USPS Result(On MNIST Domain)}
\label{fig:m2um}
\end{figure}

\begin{figure}[h]
\centering
\subfigure[Real Images]{\includegraphics[width=0.3\textwidth]{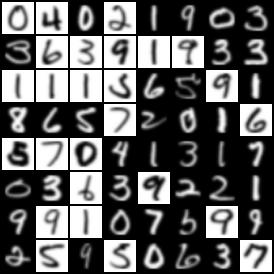}}
\subfigure[Reconstructed Images]{\includegraphics[width=0.3\textwidth]{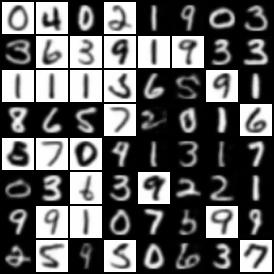}}
\subfigure[Translated Images]{\includegraphics[width=0.3\textwidth]{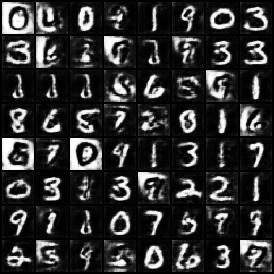}}

\caption[MNIST$\rightarrow$USPS Result(On USPS Domain)]{MNIST$\rightarrow$USPS Result(On USPS Domain)}
\label{fig:m2uu}
\end{figure}

\textbf{USPS$\rightarrow$MNIST:}

\begin{figure}[h]
\centering   
\subfigure[Real Images]{\includegraphics[width=0.3\textwidth]{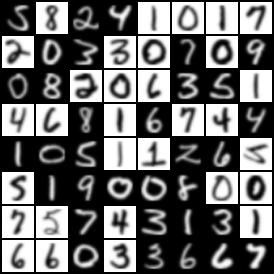}}
\subfigure[Reconstructed Images]{\includegraphics[width=0.3\textwidth]{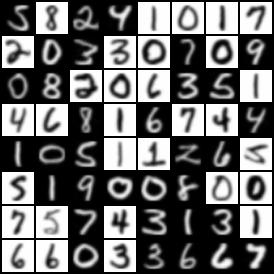}}
\subfigure[Translated Images]{\includegraphics[width=0.3\textwidth]{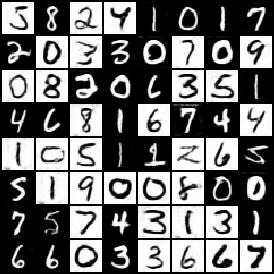}}
\caption[USPS$\rightarrow$MNIST Result(On USPS Domain)]{USPS$\rightarrow$MNIST Result(On USPS Domain)}
\label{fig:u2mu}
\end{figure}

\begin{figure}[t]
\centering
\subfigure[Real Images]{\includegraphics[width=0.3\textwidth]{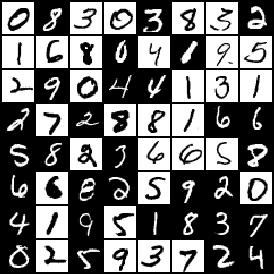}}
\subfigure[Reconstructed Images]{\includegraphics[width=0.3\textwidth]{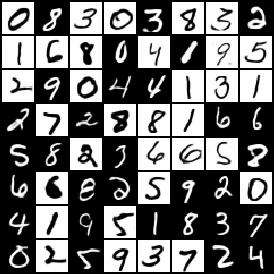}}
\subfigure[Translated Images]{\includegraphics[width=0.3\textwidth]{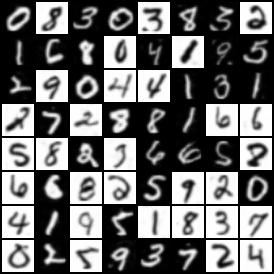}}

\caption[USPS$\rightarrow$MNIST Result(On MNIST Domain)]{USPS$\rightarrow$MNIST Result(On MNIST Domain)}
\label{fig:u2mm}
\end{figure}
\chapter{Conclusion}
In this study, I proposed a novel approach Wasserstein distance guide cross-domain learning~(WDGCDL) based on shared-latent space assumption to learn the domain invariant feature representations for domain adaptations. WDGCDL can construct a better shared-latent space by taking advantage of the gradient property of Wasserstein distance and promising generalisation bound. With the elegant share-latent space, WDGCDL can learn the invariant feature from the translated images and reconstructed images that helps to build a classifier that can be applied to the target domain.

Empirical results on two domain adaptation benchmark datasets demonstrate that WDGCDL outperforms the state-of-the-art domain adaptation framework. For the SVHN to MNIST adaptation task, I did not get good result because of the time limitation on tuning hyper-parameters. For future work, I will first tune the hyper-parameters of SVHN$\rightarrow$MNIST experiment to the best status, and then apply the framework to different task such as Image-to-Image translation. Moreover, I will investigate more information about the different label space situation and do the partial domain adaptation work.

\section{Future Work}
The future work can be divided into two parts.
\begin{enumerate}
  \item Research on tuning the SVHN$\rightarrow$MNIST experiment hyper-parameters. 
  \item Research on Image to Image translation between higher resolution domains~(e.g. Cat and Dog, day and night, apple and organ)
  \item Research on partial domain adaptation task.
  \end{enumerate}
  
 \paragraph{SVHN$\rightarrow$MNIST experiment hyper-parameters} I didn't get the best result of SVHN$\rightarrow$MNIST experiment because of the time limitation. In the future work, I will try more different hyper-parameter combinations via greedy search until get the best one.
  
 \paragraph{Image-to-Image Translation}
 The Wasserstein distance help to build a better shared-latent space across the different domains. The image to image translation task can be performed as reconstruct translated image from the shared feature representation space. Our framework more focus on the domain adaptation work on digital number domains, so that our network structures that are not very deep. To continue the research on this task, it will be useful to apply the framework on the higher resolution dataset such as imageNet or Coco datasets. With the higher resolution images, the network needs to be redesigned by adding more deep structure to capture more feature representations~(e.g. Residual Architecture).

 \paragraph{Partial Domain Adaptation}
  The framework performs the domain adaptation across different digital number datasets which share identical label spaces, (i.e. 0-9). However, that situation is no longer valid in a more realistic scenario that requires adaptation from a lager label space domain to a small label space domain. The partial domain adaptation is needed under this situation. Based on the framework and previous studies, the partial domain adaptation might be done by adding a discriminator to give less weight to the outlier class label.
\appendix
\chapter{Network Architecture Details} \label{appendix:a}
The network architecture details of the unsupervised domain adaptation tasks will be given in this section. Some abbreviations will be used for ease of presentation: n=neurons, ks=kernel size, stride=stride size. The transposed convolutional layer are represented as Deconv.

For the MNIST$\leftrightarrow$USPS experiment, the net architecture was given in Table \ref{tb:m2ueg} and Table \ref{tb:m2ud}.

\begin{table}[h]
\caption{Encoder and generator architecture for MNIST$\leftrightarrow$USPS domain adaptation.}
\label{tb:m2ueg}
\centering
\begin{tabular}{@{}clc@{}}
\toprule
Layer & Encoders                                                    & Shared? \\ \midrule
1     & Conv-(n:64, ks:5, stride:2), BatchNorm, LeakyReLU					    & No      \\
2     & Conv-(n:128, ks:5, stride:2), BatchNorm, LeakyReLU 					& Yes     \\
3     & Conv-(n:256, ks:8, stride:1), BatchNorm, LeakyReLU 					& Yes     \\
4     & Conv-(n:512, ks:8, stride:1), BatchNorm, LeakyReLU 					& Yes     \\
5     & Conv-(n:1024, ks:1, stride:1)                                          & Yes     \\
6     & FC-(n:64)                                                    & Yes     \\ \midrule
Layer & Generators                                                  & Shared? \\ \midrule
1     & FC-(N1024), ReLU                                            & Yes     \\
2     & Deconv-(n:512, ks:4, stride:2), BatchNorm, LeakyReLU                    & Yes     \\
3     & Deconv-(n:256, ks: 4, stride:2), BatchNorm, LeakyReLU                    & Yes     \\
4     & Deconv-(n:128, ks:4, stride:2), BatchNorm, LeakyReLU                    & Yes     \\
5     & Deconv-(n:64, ks:4, stride:2), BatchNorm, LeakyReLU                     & No      \\
6     & Deconv-(n:3, ks:1, stride:1), TanH                                      & No      \\ \bottomrule
\end{tabular}
\end{table}

\bigskip
\bigskip

\begin{table}[h]
\caption{Discriminator architecture for MNIST$\leftrightarrow$USPS domain adaptation.}
\label{tb:m2ud}
\centering
\begin{tabular}{@{}clc@{}}
\toprule
Layer & Discriminator                            & Shared? \\ \midrule
1     & Conv-(n:96, ks:5, stride:1), ReLU, MaxPooling-(ks:2)  & No      \\
2     & Conv-(n:192, ks:5, stride:1), ReLU, MaxPooling-(ks:2) & Yes     \\
3     & Conv-(n:384, ks:5, stride:1), ReLU, MaxPooling-(ks:2) & Yes     \\
4     & Conv-(n:768, ks:5, stride:1), ReLU, MaxPooling-(ks:2) & Yes     \\
5a    & FC-(n:1), Sigmoid                         & Yes     \\
5b    & FC-(n:10), Softmax                        & Yes     \\ \bottomrule
\end{tabular}
\end{table}
\newpage
For the SVHN $\rightarrow$ MNIST experiment, the same encoder is used, decoder and discriminator structure as the MNIST$\leftrightarrow$USPS experiment but change the latent space dimension is changed from 64 to 128. The encoder and decoder details are listed in Table \ref{tb:s2meg}

\begin{table}[h]
\caption{Encoder and generator for SVHN$\rightarrow$MNIST domain adaptation.}
\label{tb:s2meg}
\centering
\begin{tabular}{@{}clc@{}}
\toprule
Layer & Encoders                                                    & Shared? \\ \midrule
1     & Conv-(n:64, ks:5, stride:2), BatchNorm, LeakyReLU					    & No      \\
2     & Conv-(n:128, ks:5, stride:2), BatchNorm, LeakyReLU 					& Yes     \\
3     & Conv-(n:256, ks:8, stride:1), BatchNorm, LeakyReLU 					& Yes     \\
4     & Conv-(n:512, ks:8, stride:1), BatchNorm, LeakyReLU 					& Yes     \\
5     & Conv-(n:1024, ks:1, stride:1)                                          & Yes     \\
6     & FC-(n:128)                                                    & Yes     \\ \midrule
Layer & Generators                                                  & Shared? \\ \midrule
1     & FC-(n:1024), ReLU                                            & Yes     \\
2     & Deconv-(n:512, ks:4, stride:2), BatchNorm, LeakyReLU                    & Yes     \\
3     & Deconv-(n:256, ks:4, stride:2), BatchNorm, LeakyReLU                    & Yes     \\
4     & Deconv-(n:128, ks:4, stride:2), BatchNorm, LeakyReLU                    & Yes     \\
5     & Deconv-(n:64, ks:4, stride:2), BatchNorm, LeakyReLU                     & No      \\
6     & Deconv-(n:3, ks:1, stride:1), TanH                                      & No      \\ \bottomrule
\end{tabular}
\end{table}

For the 	GAN-based Network, I design a latent discriminator on latent space the details of that will be listed in Table \ref{tb:latent}

\begin{table}[h]
\caption{GAN-based Net Latent Discriminator Detail}
\label{tb:latent}
\centering
\begin{tabular}{@{}clc@{}}
\hline
Layer & Latent Discriminator & Shared? \\ \hline
1     & FC-(n:512), ReLU \hspace{5cm}     & No      \\ 
2     & FC-(n:512), ReLU      & No      \\
3     & FC-(n:512), ReLU      & No      \\
4     & FC-(n:512), ReLU      & No      \\
5     & FC-(n:2), ReLU        & No      \\ \hline
\end{tabular}
\end{table}


%

\backmatter
\bibliographystyle{ecs}
\bibliography{ECS}
\end{document}